# Image Classification Method using Dynamic Quantum Inspired Genetic Algorithm

Akhilesh Kumar Singh[a], Kirankumar R. Hiremath[b]

Email: akhileshkumarsinghcse@gmail.com[a]

Quantum Technology (IDRP), IIT Jodhpur, India[a], Department of Math, IIT Jodhpur, India[b]

**Abstract**

This study uses a metaheuristic and effective search method quantum-inspired genetic algorithm (QIGA) to choose the appropriate functionality and lower a learning dataset's dimensions, classification time, and computing cost. One way to explore a much larger solution space is by using quantum principles, namely the probabilistic nature of quantum chromosomes, represented by qubits. With the help of quantum computing principles, this algorithm seeks to improve the efficacy and efficiency of the feature selection procedure to boost performance. It also shows that without requiring enormous volumes of data, QIGA embodies both exploration and exploitation characteristics. It significantly improves classification accuracy above conventional techniques by increasing gene diversity. QIGA expresses the linear superposition of solutions using a qubit as the representation. The quantum rotation and superposition approach uses mutation and crossover processes to increase the variety and enhance population selection. QIGA's strong global search capabilities and rapid convergence make it more effective in parallel structures. This work proposes a modified QIGA known as dynamic QIGA (D-QIGA). This model is dynamically designed using the evolutionary mechanism in QIGA, allowing the optimization process to adjust to changing probabilities generated by the qubit overlay through the quantum rotation gate. The proposed algorithm employs a lengthening chromosomal technique to prevent premature convergence and local optima, providing a smooth and balanced transition between the exploration and exploitation phases. In addition, a new adaptive look-up table for rotation gates is included to improve the algorithm's optimization capabilities. Several mathematical benchmark functions and real-world restricted engineering issues are used to test D-QIGA against several well-known and cutting-edge algorithms to assess the efficacy of these concepts. The D-QIGA model provides a more accurate classification than the GA-optimized model. Compared to GA, which obtained a classification accuracy of 95%, the suggested approach performed better in classification accuracy, scoring over 99.99%.

**Keywords**- Genetic Algorithm, Quantum Genetic Algorithm, Quantum Inspired Genetic Algorithm, Quantum Mutation, Quantum Crossover

# 1. Introduction

Genetic algorithm (GA) is a metaheuristic algorithm based on natural selection belonging to the broader evolutionary algorithms class. These algorithms are based on Darwinian natural selection. Heuristic optimization techniques are widely utilized and grounded in simulated genetic processes, including mutation, crossover, and population dynamics, such as reproduction and selection (R. Lahoz-Beltra 2016). Solutions are represented in multi-arrays known as chromosomes. The chromosomes can be represent in single or multi forms. Shude Zhou and Zengqi Sun proposed single-chromosome based QIGA model (Zhou et al., 2005). The GA based-algorithm typically initiates with a randomly generated starting population of chromosomes and iteratively evolves this population in pursuit of the best solution. The limitation of this strategy is its dependence on sufficient data to train the model and enhance forecast accuracy. This constraint is consistent with the need for more precise and enough data in picture categorization.

Quantum computing emphasizes advancing computational technology grounded in the principles of quantum theory. The quantum genetic algorithm (QGA) is obtained by integrating quantum computation and GA, representing a novel probabilistic evolutionary optimization (check e.g. Han et al., 2002, Zhou et al., 2005, Zhang, 2011, and Zhou et al., 2014). Narayanan and Moore suggested a quantum evolutionary algorithm to resolve the traveling salesperson problem (Narayanan and Moore, 1996). QGA is fundamentally a variant of GA and can be utilized in any domain where traditional GA is applicable. The efficacy of QGA markedly surpasses that of the traditional GAs. The QGA exhibits a limited population size, rapid convergence rate, significant global optimization capabilities, and commendable robustness. The quantum state vector is utilized in the GA to represent genetic programming, while quantum logic gates facilitate chromosome evolution.

Han and Kim proposed a novel quantum genetic algorithm (QGA) using an optimization technique to avoid stuck in local optima and premature convergence (Han et al., 2000) and provide a solution for the knapsack problem with QIGA-based model (Han et al., 2002). The quantum rotation gate operates on qubits using superposition and updates the quantum population with specific directions and angles. It results in a program with several evaluative criteria. Furthermore, the fixed rotational angle adversely affects rapid searching and subsequent convergence (Yang et al., 2004). This study enhances the QGA with an adaptive evolution approach to address these issues. Quantum mutation, quantum population initialization, quantum fitness, and quantum crossover process are implemented to improve the performance of QIGA (Han et al., 2002 and R. Lahoz-Beltra, 2016). Quantum computation contrasts with traditional computation. It employs linear superposition to represent quantum chromosomes to facilitate quantum computation (Yang et al., 2004). Quantum computation leverages quantum mechanics approaches like entanglements and superposition for measurements of qubits through special algorithms. The capacity for parallelism is the fundamental distinction between classical (traditional) and quantum computations. In parallelism quantum computing, the problem is divided into k sub-problems. Each sub-problem is solved separately and later combined to get the solution (Zhou et al., 2014). In the probability (amplitudes) calculation, the network (device) is not in a constant condition. Conversely, it possesses a specific probability of quantum state vector corresponding to many potential states. Quantum processing utilizes the probability of amplitudes of given states (i.e., quantum states), which are squared and normalized. Consequently, the computational power of quantum systems is $\sqrt{N}$ times superior to that of classical computation. Quantum rotating gates achieve quantum transformation by using a combination of quantum gates (like Pauli gates, CNOT, Toffoli gate, etc.), and qubits can be represented as vectors with rotation (vectors states) on the Bloch sphere. Quantum computation possesses distinct features in comparison to classical computation (Zhang et al., 2011). Specific processes of these qualities can be integrated into optimization algorithms to enhance traditional optimization methods. Quantum computers utilize quantum bits, or qubits, which handle information fundamentally distinctly. Classical bits consistently represent one or zero, but quantum bits (qubits) can be described as a superposition of all possible states.

This work examines the three fundamental components of QIGA: qubit population representation, quantum rotation gate functionality, and crossover and mutation function for quantum populations. Integrating the GA with the superposition technique can significantly enhance image classification precision. Furthermore, the findings indicate that QIGA exhibits much greater efficiency than Classic GA. This advancement is attributable to QIGA's ability to achieve a more precise classification by applying the superposition principle inherent in quantum computing. Integrating the quantum tenets into evolutionary algorithms represents significant progress in feature selection methodologies. QIGA possess the potential to revolutionize healthcare diagnostics and enhance patient outcomes and classification jobs across various domains, including healthcare and treatment.

This work aims to build an algorithm capable of independently evolving dynamic quantum genetic networks for image classification, overcoming the constraints of the current conventional genetic approaches discussed here. The concept of the binary flow-diagram of QIGA programming is presented in this research to encode the structural architectures of quantum images in view of the outstanding performance measurement of quantum genetic programming in several real-life based applications. Five MNIST benchmark datasets frequently used in picture classification tasks and a real-life dataset prove the proposed method's efficacy—the suggested method's better performance over other state-of-the-art methods. computational findings demonstrates

The primarily objectives of this paper are as follows:

1. Propose a novel QIGA with a dynamic feature selection and extraction technique and choose the appropriate environment selection for image classification to improve traditional classification strategies.

2. Proposed meta-heuristic quantum inspired genetic programming with a dynamic model, where the remote archive of non-dominated and optimal solutions and the best population is achieved by a dynamic genetic quantum algorithm with quantum rotation gate with their adaptive adjustment theorems (1-2) and lookup tables-2. This approach prevents the algorithm from getting caught in a local optimum and accelerates the convergence.

The present paper has the following framework: We discuss in Section-I Introduction, Section-II literature surveys of quantum-inspired genetic algorithms with image classification and MNIST based on genetic algorithms. Section-III provides a background to quantum computing and the foundations of conventional genetic algorithms, a comparison between GA and QGA, theory-linked quantum, gate-based quantum computation, and quantum image processing. Section-IV addresses architecture and its application to the MNIST dataset as a benchmark and the suggested dynamic quantum-inspired genetic algorithm (D-QIGA). Section-V presents the outcomes gained using the architecture on several kinds of datasets and conclusion.

## 2. Literature Review

The utilization of Quantum-inspired Genetic Algorithms (QIGA) in picture classification has emerged as a significant study domain owing to their capacity to augment optimization methods and elevate model efficacy across diverse fields. Numerous studies have investigated how QGAs can enhance feature selection, augment classifier performance, and expedite convergence in image recognition tasks. This literature review analyses the contributions of several scholars in employing QGA for different picture classification issues, emphasizing both the benefits and drawbacks of their methodologies. Several academic writers are working on quantum QGAs, and their use in picture classification has aided the study, which uses the MNIST dataset. Substantial advances are made by combining quantum physics with classical approaches, increasing efficiency, accuracy, and computational performance.

J. Zhang et al. (2014) developed a multi-threshold technique with QIGA and improved histogram-based image segmentation, an important step in image categorization. The method provides adaptive rotation angle adjustment and a cooperative learning approach to improve convergence speed and solution quality. Experiments showed that the modified QIGA beats traditional algorithms in accuracy and resilience. Ranga et al. (2024) demonstrated a study of hybrid quantum neural networks (H-QNN) that used quantum computing with classical supervised learning method. Their technique produced exceptional accuracy on the MNIST dataset, outperforming previous convolutional neural network (CNN). Riaz et al. (2023) developed neural network models using Quantum Entanglement (NNQE) based on a combined Hadamard -gate with a quantum entanglement circuit. Their study findings showed advances in multi-class image classification for benchmarking datasets. (i.e. CIFAR 10 and MNIST). They demonstrated how quantum processes might improve weight selection and convergence in neural network designs, resulting in greater accuracy and performance. This model is best for noisy-based intermediate quantum computers.

Arsenii Senokosov et al. (2024) proposed two models – HQNN with a parallel quantum circuit and HQNN with a convolutional layer. The HQNN approach used for image classification reduces error with the help of reduction of image resolution via convolutional layers. Their novel embedding approach enabled efficient encoding of picture data in quantum circuits, significantly improving the performance of data reuploading models. Their research demonstrated realistic quantum implementation on benchmark datasets such as digits-MNIST, sign-MNIST, and Fashion-MNIST, offering valuable insights into the quantum encoding process. Iordanis Kerenidis et al. (2018) contributed substantially by combining slow feature analysis (SFA) with quantum genetic classification approaches. Their technique lowered dimensionality while retaining good accuracy on datasets such as MNIST. Their quantum classifier performed well, demonstrating the power of integrating conventional feature extraction approaches with quantum principles to tackle real-world picture classification problems.

Table 1 given overview of the broad and potential uses QIGAs in image categorization. These researchers proved QIGA's efficacy in improving various model parameters and emphasized its potential in machine learning and computer vision applications. QIGA has shown to be a valuable tool in image classification, optimizing hyperparameters for CNNs, QNNs, and SVMs and improving feature selection, data augmentation, picture segmentation, and even generative models such as GANs.

| Authors | Datasets | Advantage | Limitation | Contribution |
| --- | --- | --- | --- | --- |
| Lentzas et al. 2019 | MNIST, CIFAR-10 | Faster and more efficient hyperparameter optimization compared to classical algorithms. | Focused mainly on hyperparameter; no detailed comparison with advanced classical algorithms. | Explored QGA for hyperparameter tuning in machine learning. Demonstrated its potential in enhancing hyperparameter optimization classifiers. |
| Iordanis Kerenidis et al. 2018 | MNIST | High accuracy (98.5%) with polylogarithmic runtime in data dimensions and size. | Dependency on quantum hardware for scalability. | Proposed a quantum classifier combining dimensionality reduction, classification with efficiency and accuracy, comparable with classical classifiers. |
| Choe et al. 2023 | MNIST | Enhanced training efficiency and accuracy in quantum neural networks using QGA. | Requires quantum hardware for implementation; scalability issues for larger datasets. | Introduced a quantum backpropagation model optimized with QGA. suggesting the potential of QGA in quantum neural network training. |
| Creevey et al. 2023 | Synthetic datasets | Developed a GA for preparing quantum states, generating low-depth quantum | Focused on state preparation; did not directly address optimization | Introduced GASP, a GA designed to generate efficient quantum circuits for preparation of states, beneficial for |

| | | circuits for initialization. | tasks like classification. | initializing QML algorithms. |
|---|---|---|---|---|
| Arsenii Senokosov et al. 2023 | MNIST, CIFAR-10 | Improved accuracy, quantum advantage in optimization tasks | Hardware limitations for practical deployment | Developed hybrid - QNN incorporating QIGA |
| Alisson Steffens et al. 2021 | Fashion-MNIST | Reduced dimensionality, improved feature extraction | Limited adaptability to non-image datasets | Demonstrated QIGA's utility with feature selection and feature extraction. |
| Ranga et al. 2024 | MNIST | Combines quantum computing with CNN. Demonstrates effectiveness in high-dimensional classification | Focuses on binary classification. extension to multi-class classification requires further research. | Presents a novel HQNN for image classification is Combined classical computing with quantum computing to enhance performance. |
| Zhang et al. 2022 | Iris and MNIST | Proposes a robust training scheme for QNNs using a GA. Enhances accuracy distribution, leading to more reliable model performance | The impact of hyperparameters is more evident on datasets. validation on diverse datasets is necessary to confirm generalizability. | Develops a generalized training scheme with GA to improve the robustness of quantum neural networks. Validates the approach showing. Contributes to the development of more reliable QNN training methodologies. |
| Riaz et al. 2023 | MNIST and CIFAR-10 | Proposes a QGA-based neural network model for multi-class image classification. Demonstrates the potential of QGA in enhancing neural network performance. | Accuracy improvements are modest, indicating room for further optimization. performance on more complex datasets and real-world applications. | Develops a neural network model incorporating QGA for multi-class image classification. Contributes to the exploration of QGA in enhancing neural network models for image classification. |

Table 1: Summary of Literature Survey

## 3. Background

This section briefly recaps some fundamentals of GA and terminology, which will help us understand our proposed model's basic concept introduced in this section.

### 3.1 Conventional Genetic Algorithm (GA)

Genetic algorithm (GA) is a computer optimization and meta-heuristic optimization techniques that integrate several methodologies and notions derived from biological Darwinian development. They are used to tackle complicated issues by iteratively improving a population of potential solutions through the evolution process. GA solves confined and unconstrained optimization problems. These algorithms work with a collection of possible answers stored as binary digit strings or other types of data structures. A GA consists of several fundamental components, as shown in Figure 1 and described in the sub-section below.

An initial population of people, usually created randomly, is the genetic algorithm's starting point. The individuals then proceed through several cycles, referred to as generations or epochs, during which they go through crossover, mutation, and selection. These activities resemble the genetic variety, reproduction, and natural selection processes in biological evolution.

A fitness function, which measures how effectively each solution addresses the issue, is used to assess members of the existing population during the selection stage. The survival of the fittest is simulated by selecting those with better fitness scores for additional processing.

Recombination, often called crossover, is a genetic operator in which two chosen individuals exchange genetic material to produce progeny. Similar to sexual reproduction, this process creates genetically varied kids by combining the genetic material of both parents.

Mutations cause minor, arbitrary alterations to a chosen person's genetic makeup. Preserving genetic variation throughout the population makes it possible to explore various areas of the solution space.

A fresh population replaces the preceding generation after applying genetic operators. This procedure is continued for a predetermined number of generations until a termination condition is satisfied, such as achieving a target fitness level or going through a predetermined number of iterations.

Genetic algorithms search the solution space across several generations, preferring solutions with higher fitness values. By repeatedly using selection, crossover, and mutation, the algorithm can converge to an optimal or nearly optimal solution.

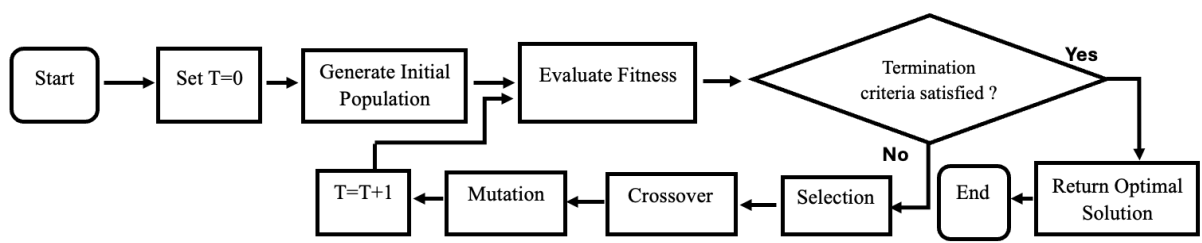

Figure 1: Overview of conventional genetic algorithm

## 3.2 Preliminaries

**Qubits:** In classical computing, the fundamental unit of information is the bit, which can denote either 0 or 1. In quantum computing, the essential unit is the qubit, which can represent $|0\rangle$ and $|1\rangle$. Their vertical representation is as:

$$|0\rangle = \begin{bmatrix} 1 \\ 0 \end{bmatrix} \qquad |1\rangle = \begin{bmatrix} 0 \\ 1 \end{bmatrix} \tag{1}$$

Pauli Dirac introduced these notation for qubits representation.

**Quantum Superposition (QS):** Superposition is a crucial property of quantum physics, permitting qubits to exist in numerous states concurrently. A traditional bit can be either 0 or 1, whereas a qubit can exist in a superposition of both states, representing both 0 and 1 simultaneously until measured. A linear combination of vectors can represent QS. Considers two vectors $|0\rangle$ and $|1\rangle$ and represent in quantum superposition-

$$|\psi\rangle = \alpha |0\rangle + \beta |1\rangle \tag{2}$$

where $\alpha$ and $\beta$ are complex numbers and should satisfy the equation-

$$|\alpha|^2 + |\beta|^2 = 1 \tag{3}$$

here, $|\alpha|^2$ and $|\beta|^2$ are amplitudes or probability of measuring both vectors state $|0\rangle$ and $|1\rangle$.

In D-QIGA, Theorems 1 and 2 are introduced to improve the rotation strategy. It helps controllability and searchability (control over to search in quantum space and qubit's states) to get optimal values of best and average fitness and reduce mutation, rotation, crossover, and selection times. Those qubit's states are closer to 90 and 0 degrees; they collapse to states '1' and '0' after the measurement process, respectively. The states of qubits are manipulated during the genetic process (mutation, crossover, and selection). After manipulation in qubits, the rotation process chooses the best state from all populations with the help of theorem 1 and 2 (using algorithm 5 and rotation table-2). In this dynamic algorithm, as 100 generations are performed, more states of qubits and qubits are collapsing and giving the best population. Every rotation is adjusted by equs. 2, 6, and 9, which is made more prominent using theorem 2.

**Theorem 1:** Consider a determinant $D = \begin{vmatrix} \alpha_i & \alpha_j \\ \beta_i & \beta_j \end{vmatrix}$; where $(\alpha_i, \beta_i)$ is represents the probability or amplitudes of some vector states with known optimal solutions. $(\alpha_j, \beta_j)$ is represents vector state's probability of finding the optimal solution.

$$f(x_i) \neq f(b_i) = \begin{cases} D \neq 0, & \text{direction of the rotation angle will be negative.} \\ D = 0, & \text{direction of the rotation angle will be either positive or negative or both} \end{cases}$$

f(x) is representing fitness of binary (0 or 1) chromosomes x and f(b) is represent best fit of individual chromosomes.

**Proof:** Consider two state vectors are $|0\rangle$ and $|1\rangle$. There are two probabilities are- $[\alpha_i, \beta_i]^T$ and $[\alpha_j, \beta_j]^T$. The first probability is known optimal solution, and second probability is unknown state to find a solution. Let angle of both probabilities ( $\alpha$ and $\beta$ ) of qubits (0 and 1) is represented by $\theta_i$ and $\theta_j$, respectively.

$$D = \begin{bmatrix} \cos\theta_i & \cos\theta_j \\ \sin\theta_i & \sin\theta_j \end{bmatrix}$$

**Case 1:**

If $D \neq 0$ $\begin{cases} 0 < |(\theta_j - \theta_i)|, & \text{direction of angle will be negative} \\ \pi < |(\theta_j - \theta_i)| < 2\pi, & \text{direction of angle will be positive} \end{cases}$

**Case 2:**

If $D = 0$ $\{ |(\theta_j - \theta_i)| < 0 \text{ or } \pi,$ direction of rotation will be positive or negetive or both

$\Delta\theta$ $(\theta_j - \theta_i)$ is deciding direction of quantum rotation gate. This theorem is used to evaluating updating transformation matrix (equ. 8).

**Theorem 2:** The best member of the previous generation's value is 'b.' The value of 'b' may or may not equal the value of chromosome 'x.' It depends upon the rotation state. We rotate the state corresponding to qubits and calculate it after measurement. The calculation can be done by combining the values of 'b' and 'x.' where 'b' and 'x' $\in [0,1]$.

$$f(x_i) \neq f(b_i) = \begin{cases} x \neq b, & \text{True or False (Postive or Negative)} \\ x = b, & \text{True or False (Positive or Negative)} \end{cases}$$

**Proof:** Let us calculate best fit of previous member and chromosomes $f(b)_i$ and $f(x)_i$ on basis of change rotation of x and b.

**Case 1:** If $f(b)_i \neq f(x)_i$ and $[(b_i =1, x_i =0) \rightarrow (b_i =0, x_i =1)]$ then

$$\Delta\theta_i = \begin{cases} (\frac{\pi}{2} - \theta_i) * C_0 \\ -\theta_i * C_0 \end{cases}$$

**Case 2:** If $f(b)_i = f(x)_i$ and $[(b_i=0, x_i=1) \rightarrow (b_i=0, x_i=0)]$ then

$$\Delta\theta_i = \begin{cases} (\frac{\pi}{2} - \theta_i) * C_1 \\ -\theta_i * C_1 \end{cases}$$

**Case 3:** if $f(b)_i$ and $f(x)_i$ may be fit or not say and $[(b_i \text{ and } x_i \text{ are same}) \rightarrow (b_i \text{ and } x_i \text{ are opposite})]$ then

$$\Delta\theta_i = \begin{cases} (\frac{\pi}{2} - \theta_i) * C_2 \\ -\theta_i * C_2 \end{cases}$$

where $C_0 = \frac{Level_{Max}}{a}$, $C_1 = \frac{Level_{Max}}{b}$, and $C_2 = \frac{Level_{Max}}{b}$

$Level_{Max}$ calculated by equ. 6.

a, b, and are constant values. The constant value for all three cases is a=1. Constant b values are taken at 20, 25, and 30, and constant c values are taken at 400, 500, and 600 for all three test cases.

$\Delta\theta_i$ is proportional to using qubit's states $\theta_i$. All three cases are calculate each rotation steps in QIGA and D-QIGA algorithms. If the rotation angle is small, test case 1 will be considered and will take a small rotation for measurement, and finding the best and average MNIST datasets is easy. If the angle is medium, then the second case, and if the angle value is very large, the third case will be performed. A large rotation angle takes more rotation for measurement and evaluates the best and average value of the MNIST dataset.

**3.3 Chromosomes-** The quantum chromosome is defined as

$$|\emptyset\rangle = a|x\rangle + b|y\rangle \qquad (4)$$

where the vectors x and y are representing states '0' and '1' respectively. Chromosomes are representing in superposition. Then the chromosomes can be representing by qubit:

$$\omega_i^p = \begin{bmatrix} a_{i,11}^p & a_{12}^p & \vdots & a_{i,21}^p & a_{i,22}^p & \vdots & ---- & \vdots & a_{i,n-1}^p & a_{i,n}^p \\ b_{i,11}^p & b_{i,12}^p & \vdots & b_{i,21}^p & b_{i,22}^p & \vdots & ---- & \vdots & b_{i,n-1}^p & b_{i,n}^p \end{bmatrix} \qquad (5)$$

The size of chromosomes is n and evaluated with generating random number ($r_{i,j} \in [0,1]$) for qubits are $[a_{i,j}^P, b_{i,j}^P]^P$, j $\in \{1,2\}$. The chromosomes qubits are collapse in these two states and generate new states of chromosomes. The population size is P, and the initialization of population by randomly.

**3.3.1 Structure and Measurement of Quantum Chromosomes-** A Quantum Chromosome is a sequence of m qubits stored in a quantum register; alternatively, it can be described as constituting a quantum register of n bits. Figure 2 illustrates the architecture of a quantum chromosome.

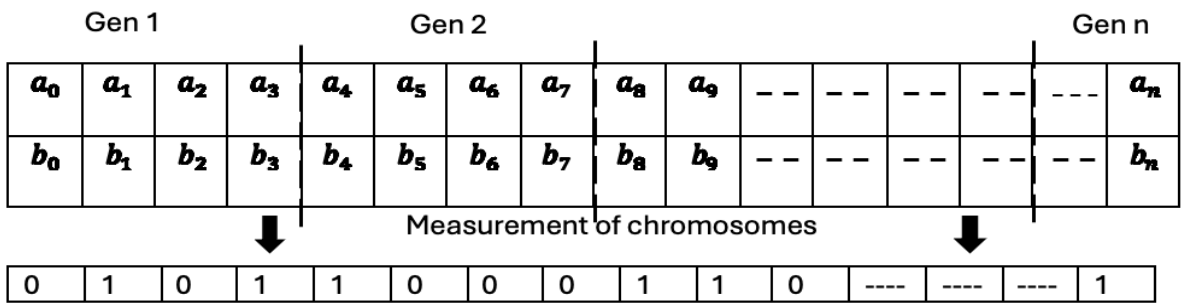

Figure 2: Sample of measurement of chromosomes

The measurement process initiates with chromosomes of minimal length for each dimension. Subsequently, at each stage, the size of the chromosomes expands by the magnitude of the interval until it attains its maximum height at the final level. The quantity of accuracy levels is determined by below equation.

$$Level_{Max} = \frac{\{Length_{Max} - Length_{Min}\}}{Interval} + 1 \qquad (6)$$

where $Length_{Max}$ = set maximum length of chromosome

$Length_{Min}$ = set initial length of chromosome

Interval = set by user (it is help to reach final length of chromosome (maximum length)).

In our experiment, we take maximum length, minimum length, and interval as 784, 1, and 1, respectively.

## 4. Methodology

This section presents our suggested combination of quantum computing techniques and classical GA. In particular, we develop a dynamic quantum-inspired genetic algorithm (D-QIGA) by partially converting a conventional GA into a quantum one.

### 4.1 Proposed Model

QIGA demonstrates effective problem-solving capabilities in specific contexts; however, it is hindered by local stagnation and excessive reliance on initial parameters. It can continue to yield high-quality responses through enhancements and collaboration with alternative methods. We present an enhanced QIGA algorithm optimized through the dynamic design of D-QIGA. Figure 3 illustrates the proposed algorithm utilized for addressing a function approximation problem. This algorithm integrates GA with a Quantum approach, as the global search capability of D-QIGA enhances the convergence speed and accuracy of QIGA. Various combination methods exist; however, we have selected to combine GA and Quantum techniques individually. This is due to our desire to quantify the extent of improvement that GA can provide for the QIGA algorithm. This method facilitates the assessment of GA's contribution to the QIGA algorithm. The proposed algorithm (D-QIGA) consists of three primary steps: QIGA, environment selection, and utilizing dynamic approach.

During the QGA step, the optimized variables are incorporated into the network as initial parameters and specific conditions must be satisfied to conclude the QGA process. This study employed optimal fitness and maximum iteration number as termination criteria. In contrast to most prior studies, this research presents a novel methodology for comparing QIGA and D-QIGA. This study is novel because previous research has not compared QIGA and D-QIGA; instead, it has only compared QIGA with the enhanced version of QIGA that incorporates QGA. D-QIGA could enhance QIGA, as the latter has overlooked the precise timing and other devotions associated with QIGA.

Furthermore, it is possible to ascertain whether a GA-combined Quantum algorithm would outperform the pure QIGA algorithm when provided with identical training time and relevant parameters. The rationale for employing distinct and separate QIGA and D-QIGA steps is now evident. The tasks outlined in the subsequent subsections are combined to form the suggested QIGA and demonstrated in figure 3.

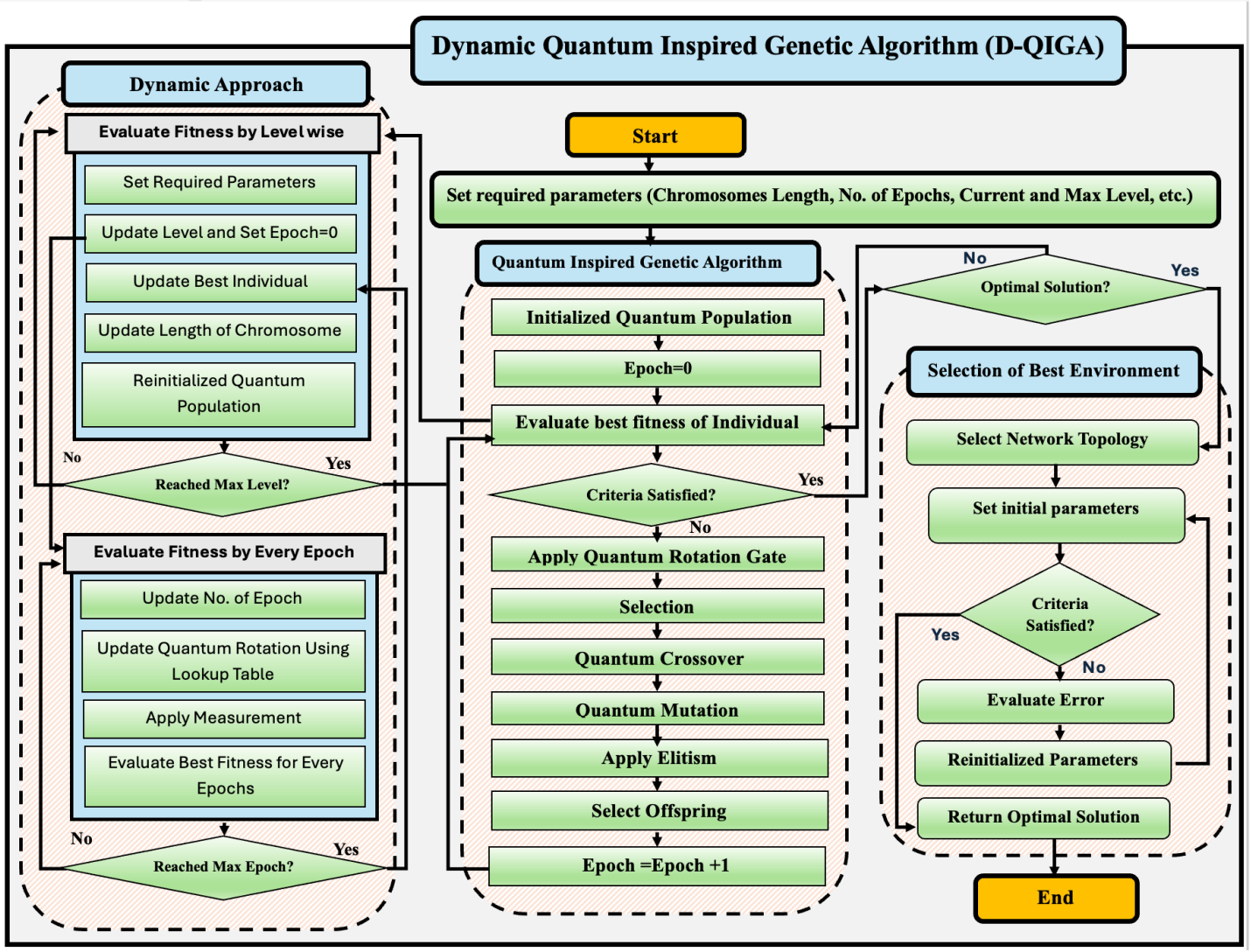

Figure 3: Proposed General Model for Dynamic Quantum Inspired Genetic Algorithm (D-QIGA)

### 4.2 Proposed Algorithms

This article proposes two novel quantum-inspired meta-heuristic algorithms: the quantum-inspired genetic algorithm (QIGA) and the quantum-inspired genetic algorithm with dynamic (D-QIGA). The pseudocode of QIGA and D-QIGA are explain in algorithm 1 and 2. Figure 3 is a description of the general framework of these suggested approaches.

The Quantum-Inspired Genetic Algorithm (QIGA) is an optimization technique that combines quantum ideas with conventional GA. The evolutionary process starts with initializing a quantum population, $P_0$, using binary image representation (Algorithm 3). Equation 10 specifies the number of generations that the algorithm will go through. In each generation, individuals' fitness is assessed using Algorithm 4, and selection (Algorithm 6) is used to designate people for the next stage depending on their fitness. Quantum rotation gates are used to update the population $P_0$, as per Algorithm 5 and Tables 2.

If certain conditions are met, quantum crossover (Algorithm 7) mixes characteristics from parent individuals to produce offspring. Quantum mutation (Algorithm 8) is then applied to offspring, adding variety while preventing premature convergence. These procedures are done on all population to increase variety and comprehensively investigate the solution space.

Following these modifications, the population's fitness is re-evaluated and ranked using Algorithm 4. The top-performing individual is preserved via the elitism process (Algorithm 9), ensuring the best solutions are preserved. Furthermore, the ideal environment matching the best individual is identified using an algorithm 11. The procedure repeats until the maximum number of generations is achieved. The program then generates the best person and environment-specific optimum solution. QIGA improves the flexibility and efficiency of classic GA by utilizing quantum-inspired approaches like rotation gates, crossover, and mutation.

---

**// Algorithm 1: Quantum Inspired Genetic Algorithm (QIGA) //**

**Input:** Size of population P
**Output:** Best individual with best environment

1. $P_0 = \emptyset$
2. $P_0 =$ Initialization quantum population with binary image (using algorithm 3)
3. for i = 1 to max $\_size\_of\_gen$ ( Using Eq. 10)
4.    $Fitness\_Val =$ Evaluating fitness (using algorithm 4)
5.    $P_1 =$ Selection ($P_0$, $Fitness\_Val$) (using algorithm 6)
6.    $Updat$ $P_0$ using quantum rotation gate ( using algorithm 5 and defined Table 2)
7.    If (criteria is not false) then
8.      for j= 1 to (P-1)
9.        $P_2 =$ quantum crossover for $P_0$ ($j$) (using algorithm 7)
10.        $P_3 =$ quantum mutation for $P_2$ ($j$) (using algorithm 8)
11.      endfor
12.    endif
13.    Evaluating and arranging individual population's fitness of $P_j$ (using algorithm 4)
14.    Select best individual with Elitism (using algorithm 9)
15.    Select best environment with best individual (using algorithm 11)
16. endfor
17. Return (best individual with best environment )

---

The Quantum-Inspired Genetic Algorithm Using Dynamic Approach **(D-QIGA)** is an innovative optimization approach that dynamically adjusts settings (length of chromosomes) during execution. The process starts with initializing parameters, such as the minimum level ($min_{Level}$), maximum level ($max_{Level}$) determined using Equation 6, and initial level ($Int_{Level}$). The default chromosomal size is set at $min_{Level}$. The method refines the population through iterations at each incremental level ($Int_{Level}$).

Quantum Rotation Gate is performed similarly to common genetic operators, but it works only on qubits and uses quantum properties. In this paper, we used the superposition of quantum states, which helps to perform genetic processes like mutation, crossovers, etc, efficiently.

In each iteration, quantum rotation gates update the quantum population Q(t) depending on its prior state Q(t−1), as specified in Table 2 (Algorithm 5). Algorithm 3 measures the quantum population to create a binary population (P(t)). This population is then subjected to quantum crossover (Algorithm 7) and mutation (Algorithm 8) to increase variety. Fitness is evaluated (Algorithm 4), and the best candidate is chosen through a selection procedure (Algorithm 6). Once a level's iterations are finished, the best individual from the current level is paired with the best individual from the previous level. The chromosomal size is increased by a predetermined interval (T), and the quantum population is reinitialized. The elitism mechanism (Algorithm 9) assures the survival of the top person while the best environment is selected (Algorithm 11). The process repeats for all levels up to $max_{Level}$. Finally, the method generates the best person and ideal environment, dynamically improving solutions via adaptive chromosomal sizing and quantum-inspired operations.

---

**//Algorithm 2: QIGA Using Dynamic Approach (D-QIGA) //**

**Input:** $min_{Level}$, $max_{Level}$, $Int_{Level}$, $No\_of\_Iteration$
**Output:** Best individual with best environment

1. Initialized D-QIGA paraments ()
2. {
3. $min_{Level}$ = 0 (by default set)
4. calculate $max_{Level}$ using Eq. 6
5. T= initialized interval
6. $Int_{Level}$ = 1 (by default set initial level)
7. t=0 (take temporary value)
8. }
9. $Chromosome_{size} = min_{Level}$
10. for $Int_{Level}$ = 1 to $max_{Level}$
11. {
12.     for i= 1 to $No\_of\_Iteration$
13.       {
14.         Updating Q(t) by using rotation gates applied to Q(t−1) as defined in Table 2. (using algorithm 6).
15.         Generate the binary population P(t) by measuring Q(t). (using algorithm 3)
16.         Apply quantum crossover for P(t). (using algorithm 7)
17.         Apply quantum mutation process for P(t). (using algorithm 8)
18.         Evaluate the fitness of P(t) and select the best individual. (using algorithm 4 and 7)
19.         Increase i by 1.
20.       }
21. endfor
22.     Reset t to 0.
23.     Increase $Int_{Level}$ by 1.
24.     $Best\_Individual\ (Int_{Level}) = Best\_Individual\ (Int_{Level} - 1)$ // Assign the top performer (best individual) from the current level to the top performer (best individual) from the prior level//
25.     $Chromosome_{size} = Chromosome_{size} + T$ // (increase size of chromosome by predefine as interval) //
26.     Reinitialize the quantum population Q(t).
27.     Select best individual with Elitism (using algorithm 9)
28.     Select best environment with best individual (using algorithm 11)
29. }
30. endfor
31. Return (Best individual with best environment)

## 4.3 Initialization of Population-
It refers to creating a new set of solutions for candidates or individuals for the starting step of quantum algorithms. It affects the quality of the solution and convergence speed in the algorithm.

Algorithm 3 creates the initial population ($P_0$) for optimization. The approach starts with an empty population and estimates the max pooling layer ($N_{PL}$) size as $\log_2 d$, where d is the size of the training dataset. For each person in the population (N), it initializes $N_{PL}^i$ to zero and creates a random number of blocks within the range ($N_{min}, N_{max}$,). Empty arrays are built to contain block and connection IDs (Block_IDs) and active nodes.

Random blocks are chosen from each block, and if they belong to the pooling layer, $N_{PL}^i$ is increased. If $N_{PL}^i$ exceeds $N_{PL}$, alternate blocks and connections are produced to maintain validity. Connections and block IDs between random blocks are computed, and the best combination of block IDs and active nodes is selected. The ideal combination ($Comb_i$) is added to the population ($P_0$). After repeating the process for all N individuals, the population ($P_0$) is wholly initialized and ready for further processing.

---

// **Algorithm 3: Initialization of population** //

**Input:** size of population N, Minimum Node = $N_{min}$, Maximum Node = $N_{max}$, size of training data set = d x d, Pooling layer= $N_{PL}$
**Output:** Initialized population $N_0$
1. $P_0 = \emptyset$
2. size of max pooling layer $N_{PL} = \log_2 d$
3. for i = 1 to N
4.     $N_{PL}^i = 0$
5.     temp = uniformly generate number between $N_{min}$ and $N_{max}$
6.     $Block\_IDs$ = create empty array (temp, temp) to store block and connection IDs
7.     $Block\_Active$ = create empty array size temp to store active nodes
8.     for j= 1 to temp
9.         rand = choose random block
10.         If (b ∈ pooling block)
11.         $N_{PL}^i = N_{PL}^i + 1$ ( increase by one)
12.         If $N_{PL}^i > N_{PL}$
13.         $rand_1$= select block remaining block list
14.         $rand_2 (m, n)$ = randomly generate connection IDs between block m and n
15.         $rand_3$ = calculate $Block\_IDs$ between two random block m and n
16.     endfor
17.     $Comb_i$ = get optimal (best ) from array $Block\_IDs$ and $Block\_Active$
18.     $P_0 = P_0 \cup Comb_i$
19. endfor
20. return $P_0$

---

## 4.4 Quantum Fitness Process-
Fitness evaluation provides a quantitative assessment to identify individuals eligible for parental roles. We test the fitness of given chromosomes by the quantum fitness function. This fitness function provides a set of fitness scores. We compared all fitness scores in the fitness set and chose a higher fitness score, which will be the optimal fitness value of the given chromosomes. Algorithm 4 illustrates the framework for fitness evaluation in QIGA. QIGA addresses image classification tasks, making classification errors the most effective criterion for determining fitness.

Each represented quantum circuit is trained on the training set $D_{Train}$, while the fitness is evaluated on a separate dataset $D_{Fitness}$. Quantum circuits often possess deep architectures, necessitating substantial computational resources and extended time for thorough training to achieve minimal classification error. A complete training process typically involves many epochs, with 100 epochs being standard for quantum circuit training. This will render the situation considerably more impractical due to the population-based QGAs involving multiple generations, wherein each individual undergoes complete training in each generation. In this method, each undergoes training for a predetermined average of epochs, specifically 100 epochs, tailored to their architectures and connection weight initialization values. Subsequently, the mean value and standard deviation of classification error are computed for each batch of $D_{Fitness}$ in the final epoch. The mean value and standard deviation of classification errors are utilized as an individual's fitness measure. A smaller mean value indicates a superior individual. When individuals being compared have identical mean values, a lower standard deviation signifies an exceptional outcome.

---

// **Algorithm 4: Evaluate fitness for quantum population** //

Input: Set of Population $P_N = \{P_0, P_1, P_2, \ldots\ldots\ldots, P_{N-1}\}$, q = number of qubits, $Total_{Epoch}$ = number of training epochs or size of batch, $D_{Train}$ = set of training dataset, $D_{Fitness}$ = set of evaluate dataset, $Measure\_Acc$ = measurement of accuracy with training epochs

Output: Fitness F with individual population

1. Construct quantum circuit
2. Evaluating training dataset with quantum circuit
3. Initialization quantum logic gate with quantum $\delta_N = \{\delta_0, \delta_1, \ldots\ldots, \delta_{N-1}\}$ where $\delta_0 = \{\theta_0, \theta_1, \ldots\ldots\ldots\ldots, \theta_{N-1}\}$
4. for each individual $\kappa$ in $P_N$
5.     i=1
6.     $every\_step_i = \frac{|D_{Fitness}|}{Total_{Epoch}}$
7.     while ( i ≤ $Measure\_Acc$)
8.         Train connecting by weights of quantum with individual $\kappa$
9.         Training quantum by gradient descent optimizer and set $\delta_N \to \delta_N'$
10.         If (i == $Measure\_Acc$)
11.             $Acc_{List} = \varnothing$
12.             for j= 1 to $every\_step$
13.                 $Acc_j$ = evaluating classification error for $D_{Fitness}$
14.                 If ( $Acc_j$ is valid for current quantum circuit )
15.                     $Acc_{List} = Acc_{List} \cup Acc_j$
16.                 j=j+1
17.             endfor
18.         update $\kappa$ from $P_N$
19.         endwhile
20.     i=i+1
21. endfor
22. return (fitness F for current population P)

**4.5 Quantum Rotation Gate-** In contrast to the GA, the QGA utilizes the probability amplitude of qubits for chromosome encoding and employs quantum rotating gates (QRG) to execute chromosomal update operations. The parental group does not decide the generation of offspring when QRG is used to perform genetic operations due to the chromosomes being in a state of superposition (entanglement). The optimal individual of the parental group and the probability amplitude of each state collectively establish it. The genetic manipulation of QGA mainly involves altering the superposition or entanglement states through QRG to modify the probability amplitude. Consequently, the development of QRG (using algorithm 5) is a critical aspect of QGA, significantly influencing the algorithm's performance.

In this method, quantum transformations are achieved using a transformation matrix, specifically employing quantum rotation gates to manipulate qubits during the mutation process. The rotation of qubits is updated using the matrix below.

$$\cup (\theta_i) = \begin{bmatrix} \cos \theta_i & -\sin \theta_i \\ \sin \theta_i & \cos \theta_i \end{bmatrix} \quad (7)$$

Then updated process is

$$\begin{bmatrix} a_i' \\ b_i' \end{bmatrix} = \cup (\theta_i) \begin{bmatrix} a_i \\ b_i \end{bmatrix} = \begin{bmatrix} \cos \theta_i & -\sin \theta_i \\ \sin \theta_i & \cos \theta_i \end{bmatrix} \begin{bmatrix} a_i \\ b_i \end{bmatrix} \quad (8)$$

Where $[\, a_i \, b_i \,]^T$ is probability amplitudes of i-th qubit of chromosome before rotating of quantum gate. After rotation and updating quantum gate through algorithm 5 and represent as $[a_i' \, b_i']^T$ for i-th qubit of chromosome. The rotation angle denoted by $\theta_i$ for i-th qubit and adjustment follow gate adjustment strategy (show in table 2), which is determined as S $(a_i, b_i)$ respectively. S $(a, b_i)\Delta\theta_i$ is defined direction of rotation (sign) and $\Delta\theta_i$ is rotation angle values (show in table 2). $\Delta\theta_i$ and S $(a_i, b_i)$ give lookup tables 2. f(x) is representing fitness of binary (0 or 1) chromosomes x and f(b) is represent best fit of individual chromosomes. The values of $\theta_i$ is adjust by gate strategy and we perform task on different values and show best fit, average fit, and random fit in table 2. the fitness values f(x) evaluated for current chromosome x and compared with optimal fit f(b) values. If f(x) is greater than f(b) then qubit adjust to direction of x with probability $(a_i, b_i)$ for ith qubit. If f(x) is less than f(b) then qubit adjust to direction of best with probability $(a_i, b_i)$ for i-th qubit.

The rotation angle is adjusted by following formula-

$$\Delta\theta_i = \theta_j - \left(\frac{\theta_j - \theta_i}{\text{Number of Repetition (per level)}}\right) * \text{Epochs} \quad (9)$$

where rotation's sign of $\theta_j$ and $\theta_i$ will be calculating with theorem 1 and 2.

| $x_i$ | $b_i$ | $f(x) \geq f(b)$ | $S(\alpha_i, \beta_i)$ | | | | | | | | | | | | $\Delta\theta_i$ | | | Apply Theorem | | |
|---|---|---|---|---|---|---|---|---|---|---|---|---|---|---|---|---|---|---|---|---|
| | | | $\alpha_i, \beta_i > 0$ | | | $\alpha_i, \beta_i < 0$ | | | $\alpha_i = 0$ | | | $\beta_i = 0$ | | | | | | | | |
| | | | T1 | T2 | T3 | T1 | T2 | T3 | T1 | T2 | T3 | T1 | T2 | T3 | T1 | T2 | T3 | T1 | T2 | T3 |
| 0 | 0 | False | 0 | - | - | 0 | + | + | 0 | ± | ± | 0 | ± | 0 | 0 | $\Delta\theta_1$ | $\Delta\theta_1$ | - | 1 and 2 | 1 and 2 |
| 0 | 0 | True | 0 | - | - | 0 | + | + | 0 | ± | ± | 0 | ± | 0 | 0 | $\Delta\theta_1$ | $\Delta\theta_1$ | - | 1 and 2 | 1 and 2 |
| 0 | 1 | False | 0 | - | + | 0 | + | + | 0 | ± | 0 | 0 | ± | ± | 0 | $\Delta\theta_3$ | $\Delta\theta_1$ | - | 1 and 2 | 1 and 2 |
| 0 | 1 | True | - | - | - | + | + | + | ± | ± | ± | 0 | ± | 0 | $\Delta\theta_2$ | $\Delta\theta_2$ | $\Delta\theta_1$ | 1 and 2 | 1 and 2 | 1 and 2 |
| 1 | 0 | False | - | + | - | + | - | - | ± | ± | ± | 0 | ± | 0 | $\Delta\theta_1$ | $\Delta\theta_1$ | $\Delta\theta_1$ | 1 and 2 | 1 and 2 | 1 and 2 |
| 1 | 0 | True | + | + | + | - | - | + | 0 | ± | 0 | ± | ± | ± | $\Delta\theta_3$ | $\Delta\theta_3$ | $\Delta\theta_1$ | 1 and 2 | 1 and 2 | 1 and 2 |
| 1 | 1 | False | + | + | + | - | - | + | 0 | ± | 0 | ± | ± | ± | $\Delta\theta_1$ | $\Delta\theta_1$ | $\Delta\theta_1$ | 1 and 2 | 1 and 2 | 1 and 2 |
| 1 | 1 | True | + | + | + | - | - | - | 0 | ± | 0 | ± | ± | ± | $\Delta\theta_3$ | $\Delta\theta_1$ | $\Delta\theta_1$ | 1 and 2 | 1 and 2 | 1 and 2 |

Table 2: Lookup Table of rotation angle with rotation direction (Test case -T1, T2, and T3) (1- Theorem 1 and 2- Theorem 2)

**4.5.1 Generation Distribution Across Various Precision Levels-** Due to QIGA's utilization of varying chromosomal sizes throughout execution, allocating a specific number of generations to each level is necessary. Shorter chromosomes necessitate fewer generations compared to those with a more significant number of genes. They are utilizing a low level of accuracy, resulting in a significantly reduced number of potential solutions compared to greater accuracy levels. Consequently, the algorithm's generations are allocated across various levels, such that when accuracy improves, the quantity of generations associated with those levels escalates. The number of repeats at each level is specified by Equation 10.

$$\text{Number of Repetition (per level)} = \frac{Level_{no}}{Max(k+1)/2} \times m \qquad (10)$$

Where $Level_{no}$ = number of levels

m = number of iterations

k = maximum length of level

m = $\sum_{i=1}^{Level_{no}}$ Repetition_Process $_i$

```
// Algorithm 5: Updating of quantum states //
Input: Population size P
Output: update quantum state
    1. i=0
    2. while (i<N)
```

```
    3.    evaluate i-th gene of best individual of population and store in Best_Amp (using
          algorithm 10)
    4.    k=0
    5.    j=0
    6.    While (j<2)
    7.        If (j! = Best_Amp )
    8.            Q[j]= c*Q        /*where c is any constant and c ∈[0,1] */
    9.            k= k+ $Q[k]^2$
    10.       endif
    11.       j=j+1
    12.   endwhile
    13.   Q[inBest_Amp] = √(1-k)
    14.   i=i+1
    15. endwhile
```

Updating Quantum States (Algorithm 5) refines a population's quantum state using the genes of the best individuals. The algorithm analyses the best gene for each person in the population and records its amplitude in Best_Amp (using Algorithms 10 ). The algorithm iterates across quantum states Q[j], scaling any states that do not meet Best_Amp with a constant c∈[0,1]. The cumulative probability (k) for these states is updated. Finally, the optimal gene's amplitude is computed as 1 - k. This procedure is repeated for all people, updating the quantum states to reflect the effect of the top-performing individual.

**4.6 Selection Process-** This process refers to choosing individuals from the quantum population to serve as parents for the creation of the next generation. In this paper, we choose the binary tournament method.

The Binary Tournament Selection algorithm (Algorithm 6) picks individuals based on their mean and parameter values. It begins by organizing the mean values of individuals in descending order ($\mu'_N$). Two random mean values ($\mu_i$ and $\mu_j$) are picked. If both are maximal, the mean ($\mu_1$, $\mu_2$), standard deviation ($Stad_1$, $Stand_2$)) and select two parameters ($\Upsilon_1$ and $\Upsilon_2$) are assessed. $\Upsilon_1$ and $\Upsilon_2$ is number of parameters are used for $\mu_i$ and $\mu_j$. The program evaluates the difference in mean values ($\mu_1 - \mu_2$) to a threshold ($\alpha$) and picks the individual with the greater mean. If the difference in parameters ($\Upsilon_1 - \Upsilon_2$) exceeds the threshold β and $Stad_1 >$ $Stand_2$, then $\mu_j$ is picked; otherwise, $\mu_i$ is chosen. If $\mu_i$ and $\mu_j$ are not the maximum, one is randomly selected. This step is repeated to complete the selection.

```
            // Algorithm 6: Binary tournament selection //
Input:  Set of individuals, α = Threshold value of mean, β = Threshold value of parameter
numbers, set of individuals with mean value $\mu_N$= {$\mu_1, \mu_2,\ldots\ldots, \mu_{N-1}$}
Output: selected only Individual
    1. Arrange set of mean value in decreasing order and store in $\mu'_N$
    2. Select two random mean values $\mu_i$ and $\mu_j$
    3. if ($\mu_i$ and $\mu_j$ are both are maximum)
    4.     $\mu_1$ and $\mu_2$ ← Evaluated mean value of $\mu_i$ and $\mu_j$
    5.     $Stad_1$ and $Stand_2$ ← Evaluated standard derivation value of $\mu_i$ and $\mu_j$
    6.     $\Upsilon_1$ and $\Upsilon_2$ ← Number of parameters used in $\mu_i$ and $\mu_j$
    7.     If (($\mu_1 - \mu_2$) > μ)
```

```
    8.         return μ_i
    9.      elseif (((Υ_1 − Υ_2)> β) and (Stad_1 > Stand_2))
   10.         return μ_j
   11.      else
   12.         return μ_i
   13.      endif
   14.   else
   15.      return random one from μ_i and μ_j
   16.   endif
   17. repeat step 2 to 16
```

**4.7 Quantum Crossover Process-** The Quantum Crossover Process (Algorithm 7) creates children by merging genes from two parent people. The newly created gene is called offspring and participates in the mutations. Two random genes ($P_{Gen,i}$ and $P_{Gen,j}$) are taken from the parent gene set ($P_{Gen}$). Gene segments ($Temp_1$ and $Temp_2$) are retrieved from randomly chosen places ($Pos_1$ and $Pos_2$) within these genes. The authenticity of the selected gene segments is verified; if so, they are assessed, edited, and united to produce a new offspring. If the chosen genes are invalid, the procedure is repeated until valid genes are discovered. The resultant offspring is returned as $Offspring_{Gen}$.

```
                    // Algorithm 7:  Quantum Crossover Process //
Input: Set of gene parent P_Gen
Output: Set of offsprings Offspring_Gen (generated by algorithm 10)
   1. P_Gen,i and P_Gen,j  ← Select rand two gen from set of gen parent P_Gen
   2. Pos_1 ← Select random position in P_Gen,i
   3. Pos_2 ← Select random position in P_Gen,j
   4. Temp_1 ← select random gen from parent P_Gen,i with position Pos_1
   5. Temp_2 ← select random gen from parent P_Gen,j with position Pos_2
   6. check validation of selected random gen Temp_1 and Temp_2
   7. if (Temp_1 and Temp_2 = Valid)
   8.    Evaluate, modified, combined and create new offspring
   9.    Offspring_Gen = new offspring
  10. else
  11.    Goto step 4
  12. endif
  13. return Offspring_Gen
```

**4.8 Quantum Mutation Process-** The Quantum Mutation Process introduces small changes in the chromosomes of individuals of a population. In this process, it selects a small string of chromosomes and changes genes by using random, swapping, reverse, etc. methods to maintain diversity and avoid getting stuck in local optima. In this paper, we are using a uniform mutation method.

Depending on the type supplied, a variety of mutation procedures are available. If the operation is Addition, a random gene (bit) is picked and placed at the specified spot. Remove selects and removes a block from the gene. If Modified is selected, a random block is changed with a new one at the given point. Swap involves selecting two genes ($P_{rand,i}$ and $P_{rand,j}$ line-13 in algorithm 8) and swapping their order if they appear in the same gene. The Inversion process involves repeatedly swapping genes between two defined places ($P_{rand,temp1}$ and $P_{rand,temp2}$, in line 18-21 in Algorithm 8) until the range is reversed. When Scramble is selected, the genes between two randomly chosen places are scrambled, resulting in a random rearrangement of the gene sequence.

When the mutation process is finished for all genes in the person, the mutated individual is returned as the output, including the newly transformed genetic material. This approach promotes greater variety and exploration of the solution space, which is essential in GA.

```
// Algorithm 8: Quantum Mutation Process//
Input: Set of parent P, Set of block n in P, Types of Operation= Addition, Remove, Modified, Swap, Inversion, Scramble, Mutation_Rate
Output: Set of Mutated offspring
   1. For each gene in individual
   2.    If (random probability <Mutation_Rate)
   3.    {
   4.          $P_{rand}$ ← Select any random position for mutation
   5.          if (Types_Operation= Addition)
   6.             Perform Addition Operation : select random gene (bit) and insert in $P_{rand}$
   7.          else if (Types_Operation =  Remove)
   8.             Perform Remove Operation : select block and remove from n
   9.          else if (Types_Operation  = Modified)
  10.             Perform Modified Operation = select random block and put into $P_{rand}$
  11.          else if (Types_Operation  = Swap)
  12.                If ($P_{rand,i}$ and $P_{rand,j}$ ∈ n)
  13.                   Perform Swap Operation = swap ( $P_{rand,i}$ and $P_{rand,j}$)
  14.          else if (Types_Operation  = Inversion)
  15.             $P_{rand,temp1}$ ← define initial position of block
  16.             $P_{rand,temp2}$ ← define last position of block
  17.            while ($P_{rand,temp1}$ < $P_{rand,temp2}$)
  18.            {
  19.                Swap ($P_{rand,temp1}$ and $P_{rand,temp2}$ )
  20.             $P_{rand,temp1}$= $P_{rand,temp1}$+1
  21.             $P_{rand,temp2}$ = $P_{rand,temp2}$ - 1
  22.            }
  23.            endwhile
  24.          else if (Types_Operation = Scramble)
  25.             shuffle (n, $P_{rand,temp1}$, $P_{rand,temp2}$)
  26.             if ( $P_{rand,temp1}$>1 and $P_{rand,temp2}$ <n)
  27.             {
  28.                 for i =  $P_{rand,temp1}$ to $P_{rand,temp2}$
  29.                    temp = i + rand ()/ ($rand\_max$/ ($P_{rand,temp2}$-i +1)
  30.                    Swap ($P_{rand,temp1}$, temp)
  31.             }
  32.       }
  33. endfor
  34. return individual
```

**4.9 Elitism Process-** Elitism selects the best individual value from every epoch and returns for the next step without going to the crossover and mutation process. It provides speed, avoids loss, and improves the performance of the genetic algorithm.

The Elitism algorithm (Algorithm 9) in QGAs is concerned with preserving the population's top performers and guaranteeing their survival throughout generations. The approach begins by picking weights for two quantum populations ($QW_i$ and $QW_j$) from a collection of quantum weights (QW). Next, the sizes of the updated quantum population ($QD_{updated}$) and $QD_j$ are specified as N and M, respectively. A temporary variable, Temp, is formed by combining a random value and the difference between $QW_i$ and $QW_j$. If an individual $D_i$'s fitness is lower than an individual $D_j$'s, weight $QW_i$ is updated by adding Temp. Otherwise, $QW_j$ updates identically. This step returns the updated quantum weights ($QW_{updated}$).

```
                // Algorithm 9: Elitism in Quantum Genetic Algorithms//
1.  Select two quantum population's weight QW_i and QW_j
2.   QW_i and QW_j ∈ QW
3.  N= size of QD_updated
4.  M= size of QD_j
5.  Temp= mod (rand ( QW_j - QW_i))
6.  if ( ø_Di < ø_Dj )
7.      QW_i = QW_i +Temp
8.  else
9.      QW_j = QW_j +Temp
10. endif
11. return set of updated QW_updated
12. for all QW_j ∈ QW_updated
13.     while ( i < N+1) do
14.        while (j < M+1) do
15.           QW_(i,j)= max _val (QW_(i,j)) + min _val (QW_(i,j) - QW_(i,j)
16.           j=j+1
17.        endwhile
18.        i=i+1
19. endwhile
20. delete all duplicated QW_updated from set of QW_updated and keep identical
21. select optimal QW_updated from set of QW_updated
22. return optimal (QW_updated)
```

**4.10 Generation of Offspring-** The objective of the generation of offspring is to increase the fitness of the entire population by employing the "best-fitted" members of the current generation to produce new offspring through processes like crossover and mutation.

The algorithm 10 for generating children comprises picking pairs of parents from the population and producing offspring via crossover processes and mutation processes. First, two random parents ($P_i$ and $P_j$) are chosen from the population ($P_N$). The algorithm will proceed if both parents are in the mating pool $\varrho$. The selected parents produce offspring using the crossover operator (as explained in Algorithm 7). Once offspring are formed, they are validated. If the children are legitimate, the mutation operator (from Algorithm 8) adds genetic variety. The children from both parents are added to the collection of offspring. The parents $P_i$ and $P_j$, are removed from the mating pool. The selection, mutation, and crossover stages are repeated until the mating pool is empty.

After iterating over the population and conducting crossover and mutation operations to pairs of parents from the mating pool, the end outcome is the set of offspring ($\zeta$). This guarantees that the population develops and adds new variants to the following generation. The pseudocode for generation of offspring is demonstrated in algorithm 10.

---

**// Algorithm 10: Generating offspring //**

Input: Set of Population $P_N = \{P_0, P_1, P_2, \ldots\ldots\ldots\ldots, P_{N-1}\}$, Mating Pool $\varrho$
Output: Set of offspring $\zeta$
1. Select two random parents $P_i$ and $P_j$
2. If ($P_i$ and $P_j \in P_N$)
3.    If ($P_i$ and $P_j \in \varrho$)
4.      generate offspring by using crossover operator (using algorithm 7)
5.      If (offspring is valid) then
6.        generate offspring by using mutation operator (using algorithm 8)
7.        $\zeta = \zeta_i \cup \zeta_j$ (store offspring)
8.    delete both parents $P_i$ and $P_j$ from mating pool
9. Repeat step 1 to 8 until mating pool are not empty.

---

**4.11 Selection of Best Environment-** In genetic algorithms, environment selection helps with network selection, optimization of necessary parameters, and error correction by allowing optimal solutions to be found and propagated, encouraging the exploration and exploitation of search spaces, and ultimately assisting in developing more efficient QGA models.

The environmental selection process explicitly addresses elitism and diversity (algorithm 9). A subset of individuals with favourable mean values is selected initially, followed by the selection of the remaining individuals using the modified binary tournament selection method outlined in Algorithm 6. The proposed QIGA and D-QIGA methods aim to enhance performance by simultaneously considering elitism and diversity through these two strategies.

The algorithm 11 for population selection using elitism and genetic operators focuses on generating a new population $P_{F+1}$ by combining the best individuals from the current population $P_F$ and the offspring $Q_F$. Initially, the number of elite individuals is determined using the elitism relative to the population size N. The top-performing individuals (with the best mean values) are selected as elites and added to $P_{F+1}$. Offspring $Q_F$ is generated using genetic operators, such as mutation and crossover. The union of $P_{F+1}$ and $Q_F$ is evaluated to identify optimal individuals for inclusion in $P_{F+1}$.

For further selection, two random individuals, $rand_1$ and $rand_2$, are chosen based on their means and standard deviations. If the mean difference ($\mu_1$ and $\mu_2$) exceeds a mean threshold value (è), or if one individual's standard deviation is smaller (indicating stability), it is selected.

Otherwise, the algorithm randomly chooses between $rand_1$ and $rand_2$. This selected individual is stored temporarily. The process repeats iteratively until the size of $P_{F+1}$ matches N, ensuring a balanced representation of elites and offspring while maintaining diversity and fitness. The final $P_{F+1}$ is returned as the next generation of the population.

```
// Algorithm 11: Selection of appropriate environment //
Input: Elitism fraction, size of population P
Output: Select population P_{F+1}
   1. temp= calculate number of elites by using Elitism with the respect to size of population N
      (using algorithm 9 )
   2. P_{F+1}= select the best (optimal) temp value, which is give best mean value in P_F ∪ Q_F
   3. /Q_F= generating offsprings with genetic operator/
   4. è= mean threshold value
   5. P_F ∪ Q_F = (P_F ∪ Q_F) + P_{F+1}
   6. Select two random values rand_1 and rand_2
   7. / rand_1 is selected individual value with maximum mean and rand_1 is another value with
      another maximum mean, where rand_1 and rand_2 ∈ ú (ú is set of mean values with
      individual values )
   8. Select standard deviation std_1 and std_2 of rand_1 and rand_2 respectively
   9. Select mean value μ_1 and μ_2 of rand_1 and rand_2 respectively
   10. If (μ_1 − μ_2 > è)
   11.    Return rand_1
   12. else if (std_1 < std_2)
   13.    return rand_1
   14. else if (std_1 < std_2)
   15.    return rand_2
   16. else
   17. return select ramdom between (rand_1 and rand_2) and store in temp_2
   18. While ( | P_{F+1}| < N) do
   19.   select temp value using step 5 to 16
   20. P_{F+1}=P_{F+1} ∪ temp_2
   21. endwhile
   22.  return P_{F+1}
```

## 5. Experimental Setup and Result Discussion

First, we used prominent feature selection techniques—listed in the preceding section—whose recall fitness scores, accuracies, loss, and error are shown to assess our proposed method's performance. The proposed model was then matched with the outcomes.

The performance results of the suggested models employing the quantum-based-selected features are then shown. The evolutionary development of the proposed quantum-inspired genetic methodology using dynamic models is illustrated since it is iteration-based. Different datasets and test cases allow the best fitness and average solutions from the feature selection models to be shown in a comparison analysis.

### 5.1 Quantum Simulators

Quantum simulators are engineered to replicate quantum working environments on traditional system. A quantum simulator replicates qubits, quantum-circuits, and environments as authentic quantum computers. To facilitate the study and construction of quantum algorithms, numerous prominent corporations and scientific research organizations have created an array of quantum simulators, such as Google Collab, VS Code, and IBM-Qiskit. Qiskit offers customers an extensive array of quantum logic gate interfaces and techniques for manipulating qubits, enabling precise control over quantum circuits.

In the proposed QIGA and D-QIGA method, we use Qiskit, VS-code and Collab for quantum circuit construction and quantum environment simulation. We summarize basic hardware and software requirements for our experimental setup in Table 3.

| Attributes | Descriptions |
|---|---|
| System | MacBook Air M2 |
| MacOS | Sequoia 15.0 |
| RAM | 16 GB (LPDDR5) |
| SSD | 512 GB |
| CPU | 8-Core (4- high performance @ 3.49 GHz, 4- energy efficiency @ 2.42 GHz) |
| GPU | 10-Core |
| GPU | 500 MHz (clock base), 1398 MHz (Boost Clock) |
| Clocks | 6400 MHz |
| Simulator | Qiskit, Google Colab, VS Code |
| Language | Python |

Table 3: Basic System Requirement

## 5.2 Simulation and Parameters Setup

The primary aim of this paper is to propose a dynamic model discovery. Enhancing the applicability of the proposed quantum-based model is intended to ensure that potential users do not need expertise in QIGA. Therefore, we establish required parameters of the proposed algorithm according to established conventions. The probabilities of crossover and mutation are detailed in Table 4. We examine three distinct test cases—1, 2, and 3—for GA, QIGA, and D-QIGA while specifying their parameters in Table 4.

| Parameters | Values for GA, QIGA, and D-QIGA | | |
|---|---|---|---|
| Test No | Test 1 | Test 2 | Test 3 |
| Size of Population | 50 | 50 | 50 |
| Number of Epochs | 100 | 100 | 100 |
| Probability of crossover | 0.2 | 0.4 | 0.6 |
| Probability of mutation | 0.5 | 0.6 | 0.8 |
| Rotation Angle | $0.001\pi$ | $0.05\pi$ | $0.08\pi$ |
| Mutation Type | Uniformly | Uniformly | Uniformly |
| Selection Type | Binary | Binary | Binary |

Table 4: Test cases with their parameters

## 5.3 Benchmark Datasets

Partially drawn from MNIST's training and testing datasets, the MNIST database includes 10,000 28x28 pixel testing and 60,000 training pictures. A similar structure is provided by the Fashion MNIST dataset, a subset of MNIST that has 10,000 testing and 60,000 training 28x28 grayscale pictures divided into 10 groups. An expansion of the NIST Special Dataset 19, the EMNIST Digits dataset consists of 240,000 training and 40,000 testing 28x28 pictures containing digits, capital and lowercase letters, and parameters structured similarly to MNIST. A 32x32 RGB image classification benchmark for ten classes, including birds and airplanes, the CIFAR-10 dataset consists of 10,000 testing and 50,000 training pictures, with equal samples in each category. Based on an expanded collection of original color pictures, the Sign-MNIST dataset contains 27,455 training and 7,172 testing 28x28 grayscale images of hand motions representing American Sign Language letters (J and Z are omitted owing to mobility limits). All dataset's specific parameters are shown in Table 5.

Table 5 delineates the attributes of the Fashion MNIST, MNIST, EMNIST Digits, CIFAR-10, and Sign MNIST datasets concerning training and test samples and the number of classes.

| Datasets | Training Samples | Testing Samples | Classes |
|---|---|---|---|
| Fashion MNIST | 60,000 | 10,000 | 10 |
| MNIST | 60,000 | 10,000 | 10 |
| DIGITS MNIST | 240,000 | 40,000 | 10 |
| CIFAR-10 | 50000 | 10000 | 10 |
| Sign MNIST | 27455 | 7172 | 10 |

Table 5: Dataset with their parameters

### 5.4 Result

This section compares the performances of quantum inspired genetic techniques utilizing the proposed featured extraction and filter-based GA selection method against standalone quantum-based learning techniques and genetic techniques, focusing solely on filter, feature extraction, and selection methods.

We compare quantum-inspired genetic techniques' best fitness, average fitness, accuracy, and loss results. This comparison utilizes the dynamic selected method selected by the proposed modified QIGA model across five high-dimensional datasets: MNIST, Fashion-MNIST, Digit-MNIST, Sign-MNIST, and CIFAR-10. The filter methods yielded optimal classification results when quantum-based learning techniques were trained using the top 10% of ranked features across five datasets. The proposed QIGA and D-QIGA methods enhanced classification accuracies and best and average fitness while improving loss. We examine three types of cases.

### 5.4.1 Comparative Analysis of Best and Average Fitness, Accuracy, and Loss

We perform three different analyses based on GA, QIGA, and D-QIGA for five types of MNIST datasets with three test cases.

Fitness Analysis of Fashion-MNIST datasets- Implement Figures 4-9 and Table 6 present the results of all three test cases for the Fashion-MNIST dataset. Figures 4 and 7 present results for the first test case; the results indicate that the best fitness levels are as follows: GA (0.9534), QIGA (0.9736), and D-QIGA (0.9989), and average fitness values are identical, with GA (0.9491), QIGA (0.9692), and D-QIGA (0.9974). Furthermore, Figures 5 and 8 present results for the second test case, indicating that the best fitness values for GA, QIGA, and D-QIGA were 0.9429, 0.9854, and 0.9991, respectively, as were the average fitness values of GA, QIGA, and D-QIGA follow: 0.9347, 0.9771, and 0.9967. Figures 6 and 9 present results for the third test case; the results indicate that the best fitness values are as follows: GA (0.9523), QIGA (0.9888), and D-QIGA (0.9992). The average fitness values are identical: GA (0.9498), QIGA (0.9857), and D-QIGA (0.9990).

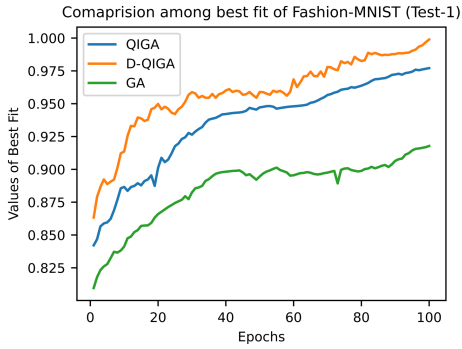 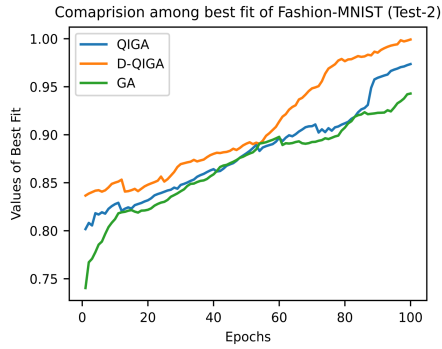 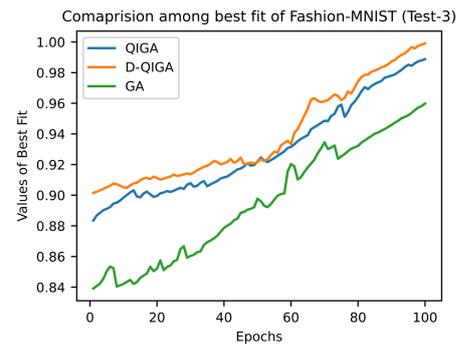

Fig. 4- Best-Fitness (Test-1)   Fig. 5- Best-Fitness (Test-2)   Fig. 6- Best-Fitness (Test-2)

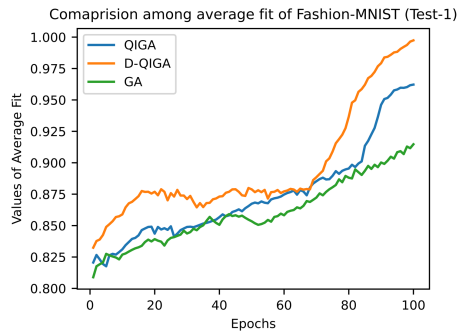 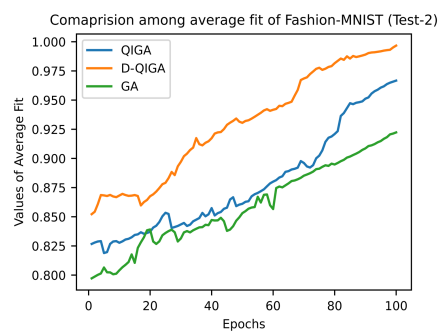 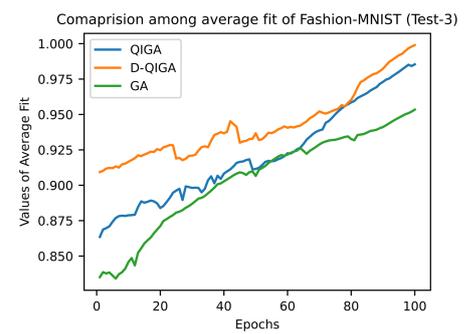

Fig. 7- Average-Fitness (Test-1) Fig. 8 - Average -Fitness (Test-2) Fig. 9 - Average -Fitness (Test-3)

Fitness Analysis of MNIST datasets- In the context of the MNIST dataset, Figures 10 and 13 and Table 6 illustrate that the best Fitness values of GA, QIGA, and D-QIGA algorithms are as follows: 0.9488, 0.9787, and 0.9986, respectively, as well as the average Fitness values of GA, QIGA, and D-QIGA algorithms are as follows: 0.9393, 0.9748, and 0.9972 respectively, were reported for the first test case. Additionally, the best and average fitness scores consistently improve by implementing the second and third test cases. Figures 11 and 14 indicate that GA, QIGA, and D-QIGA enhanced the best Fitness values of 0.9574, 0.9793, and 0.9995, respectively, as well as the average Fitness values of GA (0.9494), QIGA (0.9777), and D-QIGA (0.9987) are presented for the second test case. Improvements were made by applying the third test case and enhancing the second test case. Figures 12 and 15 indicate the best Fitness values of GA (0.9591), QIGA (0.9874), and D-QIGA (0.9988), as well as the average Fitness values of GA (0.9539), QIGA (0.9869), and D-QIGA (0.9984).

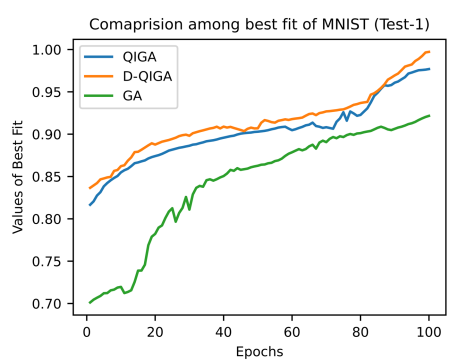 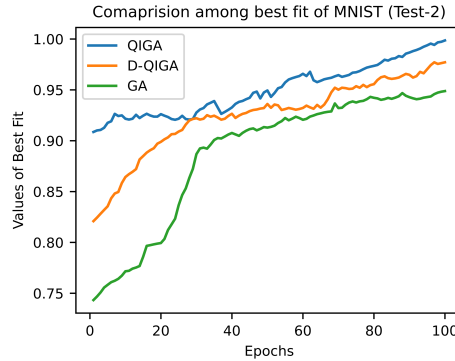 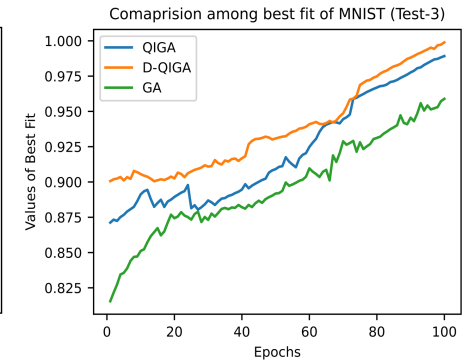

Fig. 10- Best-Fitness (Test-1)   Fig. 11- Best-Fitness (Test-2)   Fig. 12- Best-Fitness (Test-2)

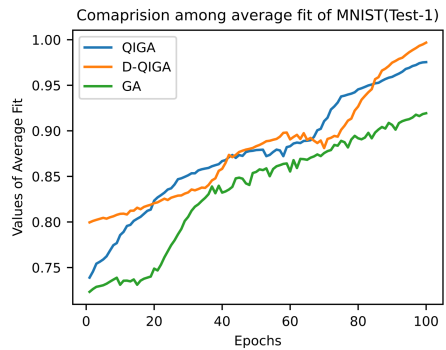
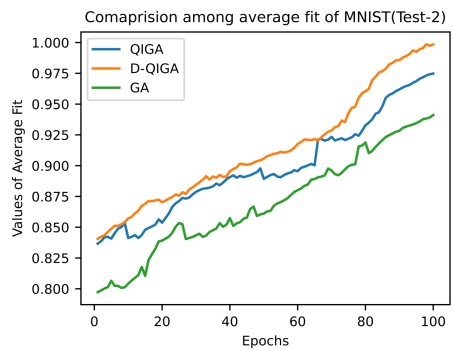
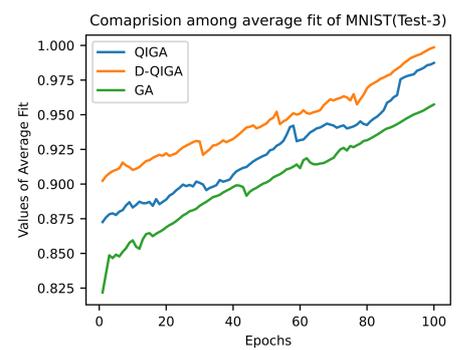

Fig. 13- Average-Fitness (Test-1)    Fig.14- Average -Fitness (Test-2)   Fig. 15- Average -Fitness (Test-3)

Fitness Analysis of Sign-MNIST datasets- In the Sign-MNIST dataset, Figure 16-21 and Table 6 illustrate that the best fitness levels achieved were 0.9705 for GA, 0.9892 for QIGA, and 0.9994 for D-QIGA. as well as the average fitness values of GA (0.9493), QIGA (0.9853), and D-QIGA (0.9992), recorded for the first test case in figure 16 and 19. Additionally, the best and average fitness scores consistently improve by implementing the second and third test cases. Applying a second test case on this dataset yields the following results: the best fitness scores are GA at 0.9418, QIGA at 0.9781, and D-QIGA at 0.9997. The average fitness scores are identical, with GA at 0.9379, QIGA at 0.9769, and D-QIGA at 0.9995, recorded in Figures 17 and 20. Furthermore, applying a third case on this dataset yields improved results, with the best fitness values recorded as follows: GA at 0.9699, QIGA at 0.9893, and D-QIGA at 0.9994. The average fitness values are identical: GA at 0.9663, QIGA at 0.9851, and D-QIGA at 0.9991 recorded in Figures 18 and 21.

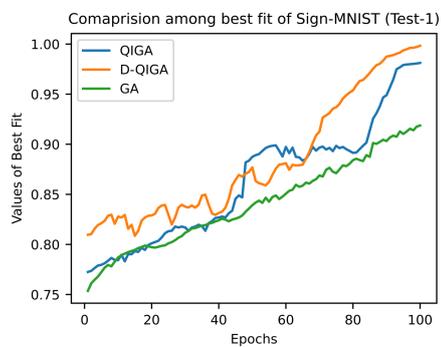
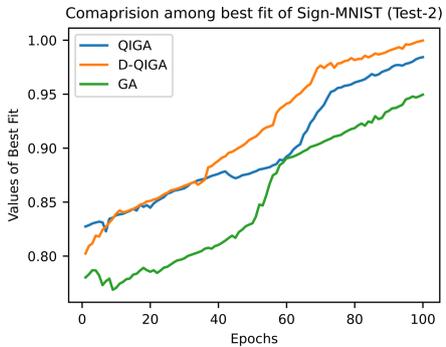
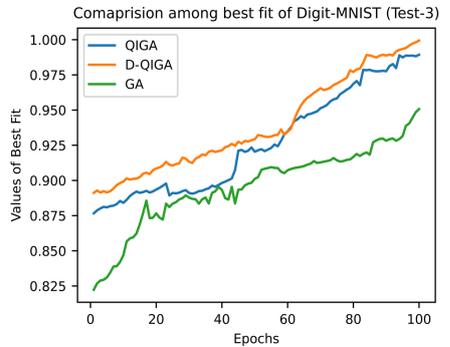

Fig. 16- Best-Fitness (Test-1)    Fig. 17- Best-Fitness (Test-2)    Fig. 18- Best-Fitness (Test-2)

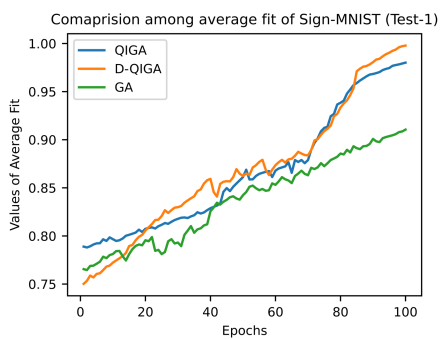
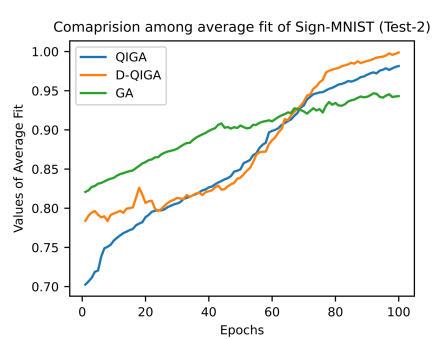
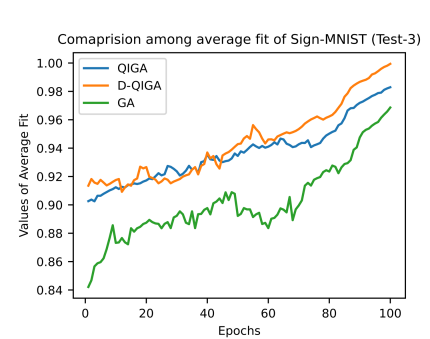

Fig. 19- Average-Fitness (Test-1) Fig. 20-Average -Fitness (Test-2) Fig. 21- Average -Fitness (Test-3)

Fitness Analysis of Digits-MNIST datasets- Figure 22-27 and Table 6 in the Digits-MNIST dataset demonstrate that the best fitness levels attained were 0.9456 for GA, 0.9995 for QIGA, and 0.9866 for D-QIGA. Average fitness scores are the same test case: GA at 0.9337, QIGA at 0.9977, and D-QIGA at 0.9856, recorded in figures 22 and 25 for the first test case. Applying a second test case on this dataset produces the following results: the best fitness scores are GA at 0.9393, QIGA at 0.9996, and D-QIGA at 0.9837. The average fitness scores are equivalent, with GA at 0.9268, QIGA at 0.9993, and D-QIGA at 0.9818, recorded in Figures 23 and 26. Implementing a third case on this dataset results in enhanced outcomes, with the best fitness values documented as follows: GA at 0.9686, QIGA at 0.9998, and D-QIGA at 0.9899 and average fitness values are equal, with GA at 0.9567, QIGA at 0.9995, and D-QIGA at 0.9874 were recorded in Figure 24 and 27.

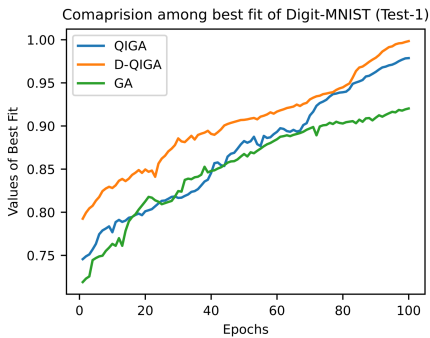
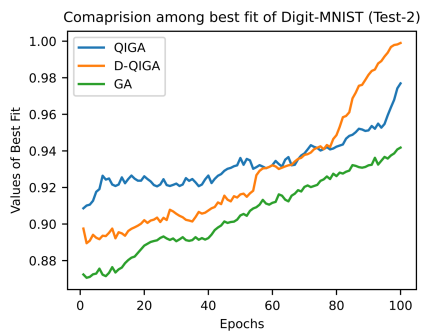
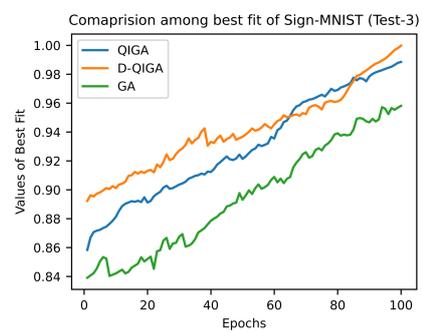

Fig. 22- Best-Fitness (Test-1)    Fig. 23- Best-Fitness (Test-2)    Fig. 24- Best-Fitness (Test-2)

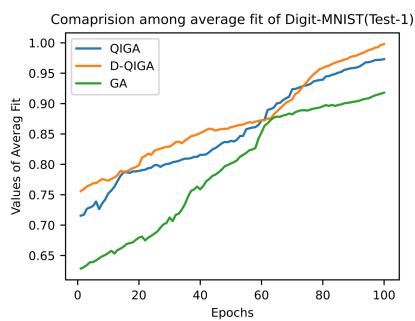
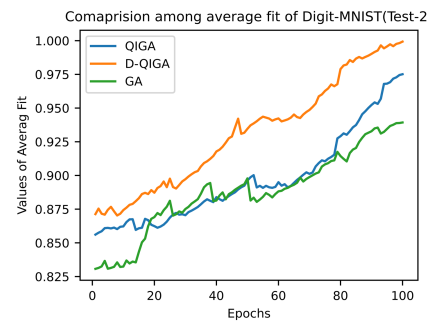
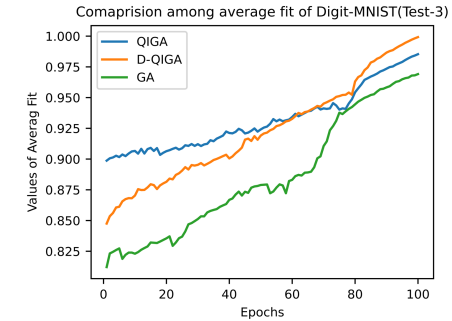

Fig. 25- Average-Fitness (Test-1) Fig. 26 - Average -Fitness (Test-2) Fig. 27 - Average -Fitness (Test-3)

Fitness Analysis of CIFAR-10 datasets- In the CIFAR-10 dataset, Figure 28-33 and Table 6 demonstrate that the best fitness levels attained were 0.9431 for GA, 0.9839 for QIGA, and 0.9994 for D-QIGA. The average fitness scores are the same in the test case: GA at 0.9387, QIGA at 0.9821, and D-QIGA at 0.9983 recorded in figures 28 and 31 for the first test case. Applying a second test case on this dataset produces the following results: the best fitness scores are GA at 0.9496, QIGA at 0.9843, and D-QIGA at 0.9997. The average fitness scores are the same: GA at 0.9485, QIGA at 0.9813, and D-QIGA at 0.9978, recorded in Figures 29 and 32. Applying a third case on this dataset results in enhanced outcomes, with the best fitness values recorded as follows: GA at 0.9686, QIGA at 0.9885, and D-QIGA at 0.9998. The average fitness values are the same: GA at 0.9581, QIGA at 0.9881, and D-QIGA at 0.9992, recorded in Figures 30 and 33.

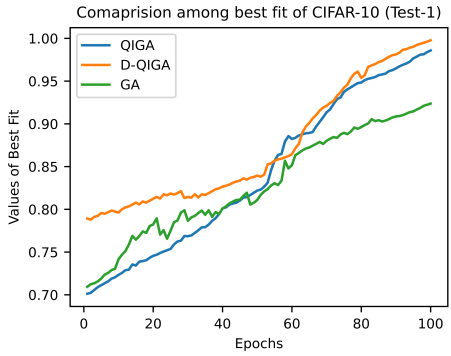
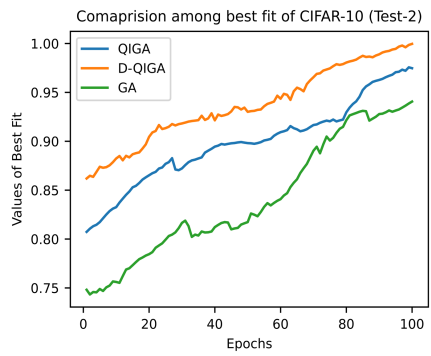
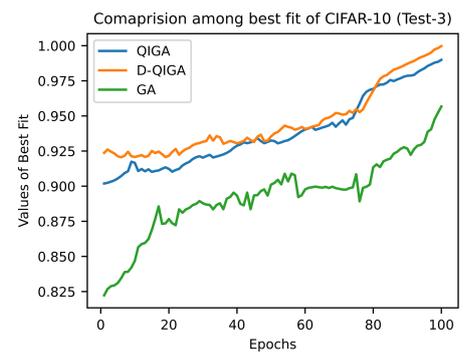

Fig. 28- Best-Fitness (Test-1)  Fig. 29- Best-Fitness (Test-2)  Fig. 30- Best-Fitness (Test-2)

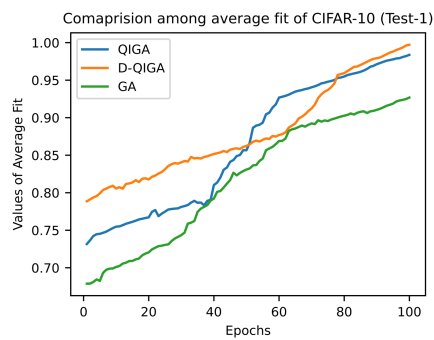
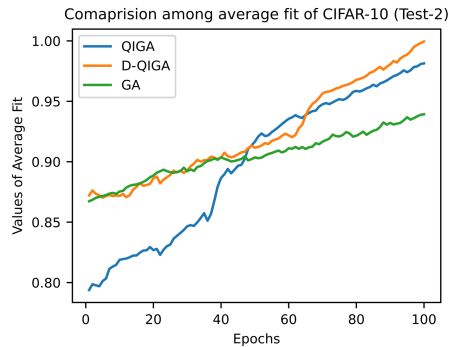
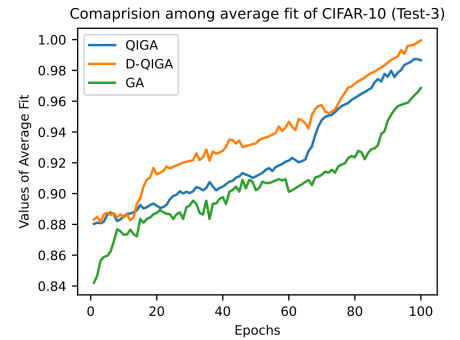

Fig. 31- Average-Fitness (Test-1)  Fig. 32 - Average -Fitness (Test-2)  Fig. 33 - Average -Fitness (Test-3)

Accuracy Analysis of datasets- Implement Figure 34 and Table 7 presents the results of the all-test case for the Fashion-MNIST dataset. The results of the first test case indicate that the classification accuracies also reflect these figures: GA (0.9598), QIGA (0.9917), and D-QIGA (0.9929). The classification accuracies of GA, QIGA, and D-QIGA were 0.9483, 0.9945, and 0.9973 for the second test case. In the third test case, the results indicate that the classification accuracies also reflect these values: GA (0.9566), QIGA (0.9972), and D-QIGA (0.9994).

In the context of the MNIST dataset, Figure 35 and Table 7 illustrate the classification accuracies for GA (0.9683), QIGA (0.9816), and D-QIGA (0.9931) reported for the first test case. Improvements were achieved by applying the second test case and enhancing the classification accuracies for GA (0.9722), QIGA (0.9933), and D-QIGA (0.9994). Improvements were made by applying the third test case and enhancing the classification accuracies for GA (0.9769), QIGA (0.9983), and D-QIGA (0.9964).

In the Sign-MNIST dataset, Figure 36 and Table 7 illustrate the classification accuracies for GA (0.9574), QIGA (0.9792), and D-QIGA (0.9926) recorded for the first test case. Additionally, the classification accuracies were similarly improved by implementing the test cases second and third. Applying a second test case on this dataset yields the following accuracy results: GA at 0.9623, QIGA at 0.9836, and D-QIGA at 0.9998. Furthermore, using a third case on this dataset yields improved results, with the classification accuracies also reflecting these figures, with GA at 0.9689, QIGA at 0.9891, and D-QIGA at 0.9890.

In the CIFAR-10 dataset, Figure 37 and Table 7 demonstrate the classification accuracies: GA at 0.9517, QIGA at 0.9692, and D-QIGA at 0.9917, recorded for the first test case. Applying a second test case on this dataset produces the following accuracy results: GA at 0.9593, QIGA at 0.9753, and D-QIGA at 0.9944. Applying a third case on this dataset results in enhanced outcomes, with the classification accuracies as follows: GA at 0.9629, QIGA at 0.9888, and D-QIGA at 0.9996.

Figure 38 and Table 7 in the Digits-MNIST dataset demonstrate that the classification accuracies are as follows: GA at 0.9703, QIGA at 0.9951, and D-QIGA at 0.9883. Applying a second test case on this dataset produces the following accuracy results: GA at 0.9662, QIGA at 0.9899, and D-QIGA at 0.9948. Implementing a third case on this dataset results in enhanced outcomes, with the classification accuracies as follows: GA at 0.9746, QIGA at 0.9988, and D-QIGA at 0.9999.

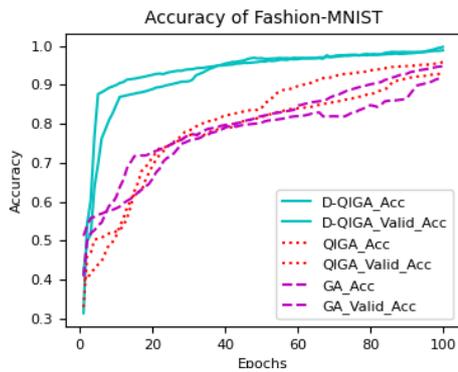 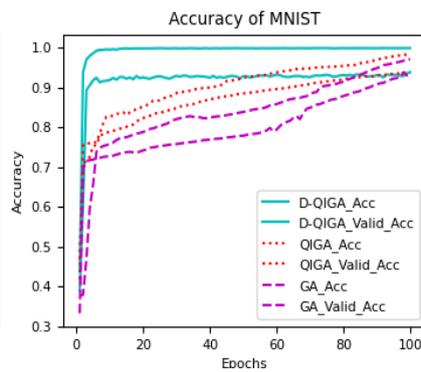 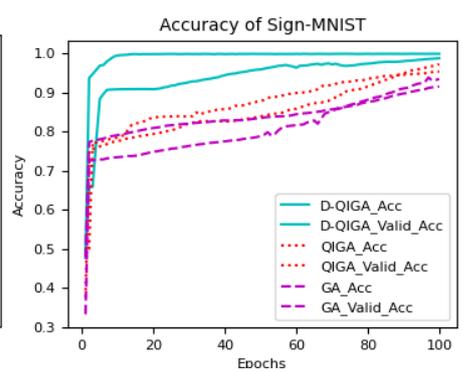

Fig. 34- Accuracy of Fashion-MNIST    Fig. 35 – Accuracy of MNIST    Fig.36- Accuracy of Sign-MNIST

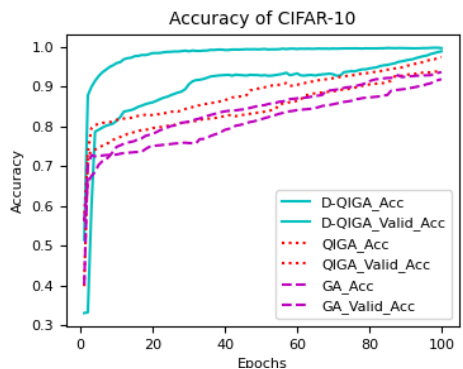 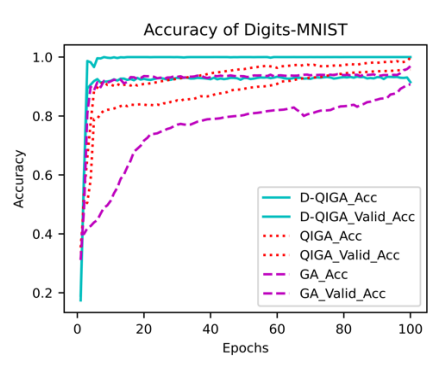

Fig. 37- Accuracy of CIFAR-10.    Fig. 38- Accuracy of Digits-MNIST

Loss Analysis of datasets- Implement Figure 39 and Table 7 present the results of the all-test case for the Fashion-MNIST dataset. The results of the first test case indicate that the loss values of GA, QIGA, and D-QIGA were observed as follows: 0.5878, 0.1123, and 0.0583. The loss values of GA, QIGA, and D-QIGA were observed as follows: 0.6372, 0.1092, and 0.0517 after applying the second test case. Implementing a third case on this dataset results in enhanced outcomes, with the loss of GA, QIGA, and D-QIGA being 0.5991, 0.0608, and 0.0412, respectively.

In the context of the MNIST dataset, Figure 40 and Table 7 illustrate the loss metrics for GA (0.4618), QIGA (0.0924), and D-QIGA (0.0641) reported for the first test case. Improvements were achieved by applying the second test case and enhancing the first test case, and the loss metrics were found for GA (0.4181), QIGA (0.0637), and D-QIGA (0.0396). Improvements were made by applying the third test case and enhancing the second test case, and the loss metrics were calculated for GA (0.3671), QIGA (0.0479), and D-QIGA (0.0426).

In the Sign-MNIST dataset, Figure 41 and Table 7 illustrate that the losses achieved were for GA (0.5506), QIGA (0.3233), and D-QIGA (0.0381), recorded for the first test case. The losses were similarly improved by implementing the second and third test cases. Applying a second test case on this dataset yields the following results: GA (0.5165), QIGA (0.0747), and D-QIGA (0.0202). Furthermore, applying a third case on this dataset yields improved results, with the loss metrics for GA (0.4871), QIGA (0.0582), and D-QIGA (0.1288).

In the CIFAR-10 dataset, Figure 42 and Table 7 demonstrate that the loss metrics were calculated for GA (0.4419), QIGA (0.1035), and D-QIGA (0.0911), recorded for the first test case. Applying a second test case on this dataset produces the following results: the loss metrics for GA (0.4276), QIGA (0.0837), and D-QIGA (0.0686). Applying a third case on this dataset results in enhanced outcomes, with the loss metrics for GA (0.4041), QIGA (0.0622), and D-QIGA (0.0206).

Figure 43 and Table 7 in the Digits-MNIST dataset demonstrate that the losses attained were GA (0.3984), QIGA (0.0898), and D-QIGA (0.1274) for the first test case. Applying a second test case on this dataset produces the following results: loss metrics for GA (0.4386), QIGA (0.1088), and D-QIGA (0.0595). Implementing a third case on this dataset results in enhanced outcomes, with the GA (0.3517), QIGA (0.0387), and D-QIGA (0.0104).

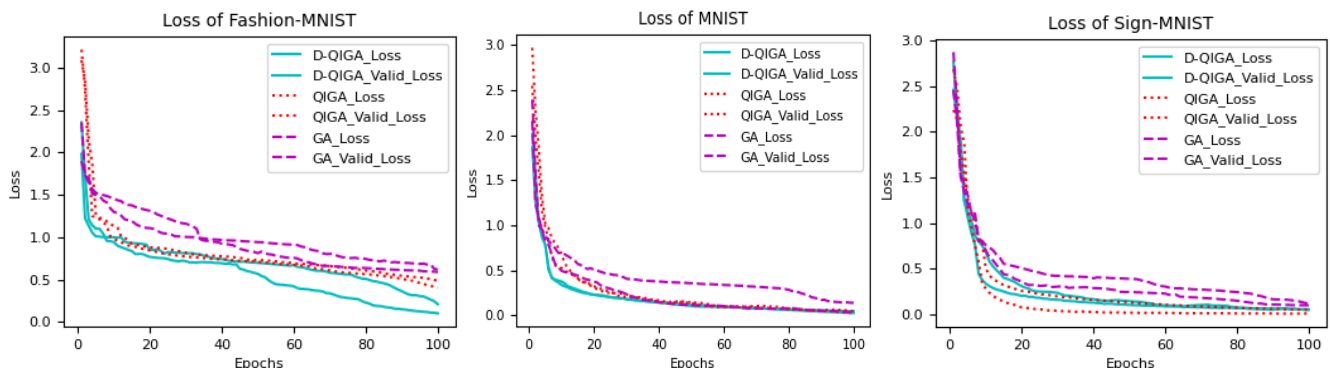

Fig. 39- Loss of Fashion-MNIST    Fig. 40 - Loss of MNIST Fig.    41- Loss of Sign-MNIST

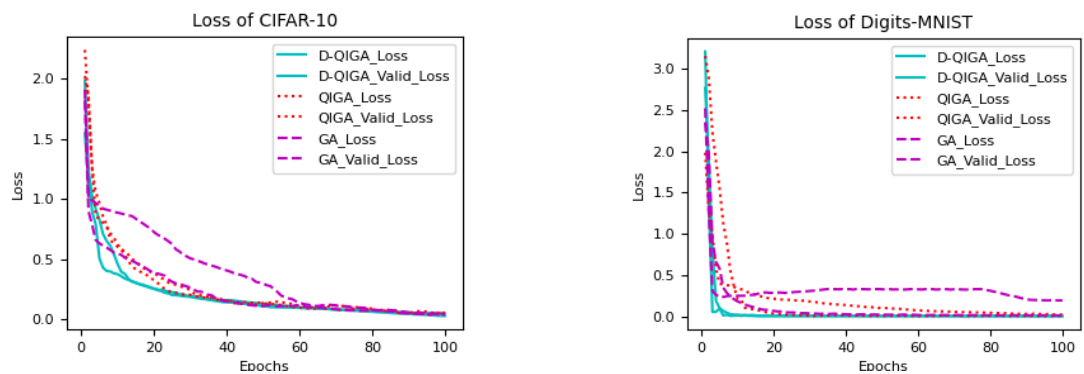

Fig. 42- Loss of CIFAR-10                    Fig. 43- Loss of Digits-MNIST

Tables 6 and 7 present the classification accuracy alongside performance metrics such as Recall and Precision and the best and average fitness and loss measures for GA, QIGA, and D-QIGA, both before and after implementing the proposed dynamic selection QIGA-based methods.

| Database | Test No. | GA | | QIGA | | D-QIGA | |
|---|---|---|---|---|---|---|---|
| | | Best Fit | Avg-Fit | Best Fit | Avg Fit | Best Fit | Avg Fit |
| **Fashion-MNIST** | 1 | 0.9534 | 0.9491 | 0.9736 | 0.9692 | 0.9989 | 0.9974 |
| | 2 | 0.9429 | 0.9347 | 0.9854 | 0.9771 | 0.9991 | 0.9967 |
| | 3 | 0.9523 | 0.9498 | 0.9888 | 0.9857 | 0.9992 | 0.9990 |
| **MNIST** | 1 | 0.9488 | 0.9393 | 0.9787 | 0.9748 | 0.9986 | 0.9972 |
| | 2 | 0.9574 | 0.9494 | 0.9793 | 0.9777 | 0.9995 | 0.9987 |
| | 3 | 0.9591 | 0.9539 | 0.9874 | 0.9869 | 0.9988 | 0.9984 |
| **Signs-MNIST** | 1 | 0.9507 | 0.9493 | 0.9892 | 0.9853 | 0.9994 | 0.9992 |
| | 2 | 0.9418 | 0.9379 | 0.9781 | 0.9769 | 0.9997 | 0.9995 |
| | 3 | 0.9699 | 0.9663 | 0.9893 | 0.9851 | 0.9994 | 0.9991 |
| **Digits-MNIST** | 1 | 0.9456 | 0.9337 | 0.9866 | 0.9856 | 0.9995 | 0.9977 |
| | 2 | 0.9393 | 0.9268 | 0.9837 | 0.9818 | 0.9996 | 0.9993 |
| | 3 | 0.9686 | 0.9567 | 0.9899 | 0.9874 | 0.9998 | 0.9995 |
| **CIFAR-10** | 1 | 0.9431 | 0.9387 | 0.9839 | 0.9821 | 0.9994 | 0.9983 |
| | 2 | 0.9496 | 0.9485 | 0.9843 | 0.9813 | 0.9997 | 0.9978 |
| | 3 | 0.9686 | 0.9581 | 0.9885 | 0.9881 | 0.9998 | 0.9992 |

Table 6: Best and Average Fitness of Fashion- MNIST, MNIST, Signs- MNIST, Digits-MNIST, and CIFAR-10

| Model | Test case | Evaluation | Fashion-MNIST | MNIST | CIFAR-10 | Digits-MNIST | Sign-MNIST |
|---|---|---|---|---|---|---|---|
| **GA** | 1 | Accuracy | 0.9598 | 0.9683 | 0.9517 | 0.9703 | 0.9574 |
| | | Loss | 0.5878 | 0.4618 | 0.4419 | 0.3984 | 0.5506 |
| | 2 | Accuracy | 0.9483 | 0.9722 | 0.9593 | 0.9662 | 0.9623 |
| | | Loss | 0.6372 | 0.4181 | 0.4276 | 0.4386 | 0.5165 |
| | 3 | Accuracy | 0.9566 | 0.9769 | 0.9629 | 0.9746 | 0.9689 |
| | | Loss | 0.5991 | 0.3671 | 0.4041 | 0.3517 | 0.4871 |
| **QIGA** | 1 | Accuracy | 0.9917 | 0.9816 | 0.9692 | 0.9951 | 0.9792 |
| | | Loss | 0.1123 | 0.0924 | 0.1035 | 0.0898 | 0.3233 |
| | 2 | Accuracy | 0.9945 | 0.9933 | 0.9753 | 0.9899 | 0.9836 |
| | | Loss | 0.1092 | 0.0637 | 0.0837 | 0.1088 | 0.0747 |
| | 3 | Accuracy | 0.9972 | 0.9983 | 0.9888 | 0.9988 | 0.9891 |
| | | Loss | 0.0608 | 0.0479 | 0.0622 | 0.0387 | 0.0582 |
| **D-QIGA** | 1 | Accuracy | 0.9929 | 0.9931 | 0.9917 | 0.9883 | 0.9926 |
| | | Loss | 0.0583 | 0.0641 | 0.0911 | 0.1274 | 0.0381 |
| | 2 | Accuracy | 0.9973 | 0.9994 | 0.9944 | 0.9948 | 0.9998 |
| | | Loss | 0.0517 | 0.0396 | 0.0686 | 0.0595 | 0.0202 |
| | 3 | Accuracy | 0.9994 | 0.9964 | 0.9996 | 0.9999 | 0.9890 |
| | | Loss | 0.0412 | 0.0426 | 0.0206 | 0.0104 | 0.1288 |

Table 7: Accuracy and Loss of Fashion- MNIST, MNIST, Signs- MNIST, Digits-MNIST, and CIFAR-10

### 5.4.2 Comparative Analysis of Determining Rotating Angle Direction, Mutation, and Crossover Process

Table 8 shows the statistical result of 100 generations evolutionary process (GA) and quantum-inspired genetic algorithm (QIGA and D-QIGA)) of a given time of optimal, worst, and average of rotation of angle's direction. The genetic process slows down between 40 and 70 generations. Between 60 and 100 generations yields the ideal answer.

This work investigates three scenarios; the highest and minimum rotation angles of the three cases are correspondingly (a) (0.001π, 0.005π) (b) (0.05π, 0.08π ) and (c) (0.001π, 0.08π). All cases use a quantum rotation angle algorithm (algorithm 5) for GA, QIGA and D-QIGA. For the three cases, 100 trials were taken; the experimental findings were displayed in Table 8; so, the rotating strategy was chosen for the three cases in this study.

Based on the angle rotation case ((a) (0.001π, 0.005π) (b) (0.05π, 0.08π ) and (c) (0.001π, 0.08π)) of the quantum genetic algorithm, the quantum mutation process is added with a probability of mutation 0.5, 0.6, and 0.8. By running the quantum mutation algorithm (algorithm 4) 100 times, Table 8 presents the experiment's statistical (optimal, worst, and average time) findings for GA, QIGA, and D-QIGA. We can thus conclude once more that the speed of development is enhanced. Applying the quantum mutation method in five MNIST datasets allows us to get the optimal solution with 100 generations. Once more, the speed of evolution is enhanced; convergence times likewise increase.

The quantum crossover process is incorporated with a probability of mutation 0.2, 0.4, and 0.6 based on the angle rotation case ((a) (0.001π, 0.005π) (b) (0.05π, 0.08π ) and (c) (0.001π, 0.08π)) and quantum mutation 0.5, 0.6, and 0.8 of the quantum genetic algorithm. Table 8 shows the experiment's statistical results (optimal, worst, and average time) by performing the quantum crossover method (algorithm 6) 100 times for GA, QIGA, and D-QIGA. Once more, we can conclude that development's pace is accelerated. The quantum crossover approach in five-MNIST databases lets us find the best solution with 100 generations. Once more, the development speed is improved; convergence times also increase. Please note that experiments may vary in hardware configuration and internet speed for dataset training and feature extraction processes.

Table 8 shows the experiment's statistical results (optimal, worst, and average time) by performing the rotation strategy (algorithm 5), quantum mutation algorithm (algorithm 8), and quantum crossover method (algorithm 7) 100 times for GA, QIGA, and D-QIGA. Once more, the development speed is improved; convergence times also increase. Please note that experiments may vary in hardware configuration and internet speed for dataset training and feature extraction processes.

| Test Case | Rotating Angle Strategy | Model | Rotation Strategy | | | Mutation in QGA | | | Crossover in QGA | | |
|---|---|---|---|---|---|---|---|---|---|---|---|
| | | | Optimal | worst | Average | Optimal | worst | Average | Optimal | worst | Average |
| 1 | (a) | GA | 25.7601 | 28.6613 | 26.7345 | 25.9988 | 29.2387 | 26.9979 | 26.5643 | 29.8112 | 27.0267 |
| | | QIGA | 14.6776 | 18.7093 | 15.4712 | 15.4327 | 18.9907 | 15.7868 | 15.9896 | 19.2355 | 16.3423 |
| | | D-QIGA | 12.5478 | 16.5907 | 13.7345 | 12.8601 | 16.9021 | 13.9911 | 13.2345 | 17.3659 | 13.9125 |
| | (b) | GA | 14.8522 | 15.8691 | 14.8566 | 15.2379 | 16.2148 | 15.8769 | 15.8124 | 16.6102 | 16.2768 |
| | | QIGA | 9.4286 | 10.9327 | 9.7136 | 9.9843 | 11.5453 | 10.6965 | 10.9666 | 12.8521 | 11.0011 |
| | | D-QIGA | 8.4533 | 9.8565 | 8.7021 | 8.5783 | 10.0066 | 8.9672 | 9.2432 | 10.8995 | 9.7756 |
| | (c) | GA | 22.7589 | 25.7742 | 23.0684 | 23.0453 | 26.1482 | 23.8461 | 24.0122 | 26.7333 | 24.8726 |
| | | QIGA | 17.9264 | 19.9664 | 17.9895 | 18.1234 | 20.6523 | 18.3897 | 18.7392 | 22.2351 | 19.0326 |
|   |     | D-QIGA | 13.7369 | 14.7521 | 13.9682 | 14.2439 | 15.9859 | 14.6999 | 15.1218 | 16.1122 | 15.9001 |
| 2 | (a) | GA     | 19.5843 | 21.9678 | 19.9579 | 19.7528 | 21.8839 | 19.9977 | 20.9893 | 23.9872 | 21.5864 |
|   |     | QIGA   | 12.8947 | 13.8672 | 12.9981 | 12.8982 | 13.9695 | 13.1127 | 13.7866 | 15.3587 | 13.9925 |
|   |     | D-QIGA | 9.9678  | 10.9911 | 10.1581 | 10.8276 | 12.0163 | 11.0062 | 11.7114 | 12.8427 | 12.1735 |
|   | (b) | GA     | 14.8525 | 15.9468 | 14.8599 | 14.6763 | 16.1932 | 14.9864 | 15.9869 | 16.7488 | 15.9736 |
|   |     | QIGA   | 9.7165  | 10.8945 | 10.1143 | 9.9816  | 11.0745 | 10.2192 | 9.9988  | 10.9893 | 10.1534 |
|   |     | D-QIGA | 5.9437  | 6.8773  | 6.1875  | 6.1986  | 7.5188  | 6.4971  | 7.1834  | 8.2474  | 7.5147  |
|   | (c) | GA     | 21.7491 | 23.6745 | 22.0444 | 22.8256 | 24.3795 | 23.0376 | 23.7631 | 25.9867 | 23.8943 |
|   |     | QIGA   | 19.9854 | 20.9682 | 20.2259 | 20.4127 | 21.7886 | 20.6991 | 21.3897 | 22.8934 | 21.9275 |
|   |     | D-QIGA | 8.9783  | 9.9674  | 9.1372  | 9.8945  | 10.8679 | 9.4765  | 9.8948  | 11.9271 | 10.3791 |
| 3 | (a) | GA     | 20.9874 | 22.8962 | 21.7856 | 21.9873 | 23.9856 | 22.0898 | 22.7815 | 24.8279 | 23.0862 |
|   |     | QIGA   | 10.6843 | 11.8875 | 10.3654 | 10.8862 | 11.9327 | 11.0131 | 11.9986 | 12.7892 | 12.0163 |
|   |     | D-QIGA | 8.4691  | 9.8892  | 8.9145  | 8.9333  | 10.2537 | 9.3569  | 9.9054  | 10.9863 | 10.1212 |
|   | (b) | GA     | 22.6291 | 23.8673 | 22.9852 | 23.4798 | 25.9539 | 23.8987 | 23.9895 | 26.9199 | 24.6874 |
|   |     | QIGA   | 10.7877 | 11.8334 | 11.0302 | 10.9012 | 12.0618 | 11.1788 | 11.8951 | 12.9989 | 12.1792 |
|   |     | D-QIGA | 7.8976  | 8.8218  | 8.0347  | 7.6971  | 8.5629  | 7.9682  | 8.3743  | 9.9754  | 8.8669  |
|   | (c) | GA     | 20.5633 | 21.9487 | 20.8722 | 21.8986 | 23.0843 | 22.0761 | 23.1985 | 25.1493 | 23.5856 |
|   |     | QIGA   | 14.6937 | 15.9891 | 14.9031 | 14.3237 | 15.3869 | 14.8792 | 15.8796 | 16.7929 | 16.2143 |
|   |     | D-QIGA | 9.5489  | 10.5867 | 9.9785  | 9.8955  | 10.8788 | 10.1043 | 10.9984 | 12.4867 | 11.8779 |

Table 8: Optimal, Worst, and Average time (s) of direction rotation of angle, mutation, and crossover for MNIST-Datasets

## 5.5 Complexity Analysis

The worst-case scenario time complexity analysis of our proposed model D-QIGA is defined in the steps below.

1. In the D-QIGA model, Worst Time Complexity ($T_{Complex}$) is used to prepare the initialization of chromosomes with population size P and individual length $Ind_{Length}$. (where length $Ind_{Length}$ is measured by maximum intensity of image as $Ind_{Length}=Int_{img}$ x $Int_{img}$.) is O(P x $Ind_{Length}$) or O(P x$Int_{img}$ x $Int_{img}$).

2. In the next step, each chromosome is prepared with quantum state population matrix Q concerning P, and $T_{Complex}$ is O(P x $Ind_{Length}$).

3. Evaluate threshold value with specific parameters (probability parameters) and capacity, then $T_{Complex}$ is O(P x $Ind_{Length}$).

4. The $T_{Complex}$ of measure feature selection and extraction (quality) of the image with quantum encoded of each pixel (image) is O(P x $Ind_{Length}$).

5. The $T_{Complex}$ evaluates the fitness of each chromosome is O(P).

6. The $T_{Complex}$ of the quantum rotation gate process used on the quantum population to give a new quantum state with the best chromosome is O(P x $Ind_{Length}$).

7. The $T_{Complex}$ of the quantum selection process, quantum mutation, and quantum crossover is O(P x $Ind_{Length}$).

8. The overall time complexity ($T_{Complex}$) of our proposed model D-QIGA is O(P x $Ind_{Length}$ x $Iteration_{max}$).

**5.6 Comparative Analysis-** Combining GA with Quantum-based dynamic approach improved classification accuracy ratios for Fashion-MNIST, MNIST, Sign-MNIST, Digits-MNIST, and CIFAR-10 using 100 epochs of experimentation. QIGA and D-QIGA, with 100 epochs, produced the most accurate categorization results. Table 7 shows how the suggested approach can accurately classify all five databases. We compared our proposed model with other author's models and show the comparison in Table 9 below.

| Reference | Used Datasets | Used Approach | Complexity Analysis | Accuracy (%) |
|---|---|---|---|---|
| W. Ye et al., 2020 | MNIST | QIEA | No | 98.97 |
| Andrés et al., 2023 | MNIST | FCNN and QCNN | No | 98 |
| Y. Sun et al., 2020 | CIFAR-10 | CNN-GA | No | 96.78 |
| Liu et al., 2023 | CIFAR100, CIFAR10, Fashion, MRB, MRI MRDBI, MB, and MRD. | Leader–Follower Evolution Mechanism with Multi-Object Genetic Programming (LF-MOGP) | No | 98.07 |
| Li et al., 2023 | MNIST | Evolutionary Quantum Neural Architecture | No | 98.99 |
| Proposed Model | Fashion-MNIST | QIGA and QIGA-based with Dynamic approach (D-QIGA) | Yes | 99.72 and 99.94 |
| | MNIST | | | 99.83 and 99.94 |
| | Sign-MNIST | | | 98.91 and 99.98 |
| | Digits-MNIST | | | 99.88 and 99.99 |
| | CIFAR-10 | | | 98.88 and 99.96 |

Table 9: Comparison of proposed model with other models

**5.7 Conclusion**

In this paper, we proposed two algorithms- QIGA and D-QIGA. D-QIGA is a modification of the QGA algorithm. These algorithms are introduced for image classification for MNIST datasets. Both algorithms concentrate on facilitating the transition between the phases, as mentioned earlier, in addition to exploration and exploitation skills. An adaptive look-up table and a lengthening chromosomes technique are used to improve these algorithms. By balancing the exploration and extraction phases and facilitating a seamless transition between them, the lengthening chromosomes strategy helps to avoid premature convergence of the algorithm and trapping it in local optima. It also gradually clarifies the search space and delays the convergence to high-precision solutions to the ending generations. The D-QIGA algorithm is used for experimental testing and can suggested for medium-sized datasets. In three cases, D-QIGA performed better than the comparison algorithms GA and QIGA, particularly for MNIST datasets.

Future research can use D-QIGA to solve additional optimization problems, including feature selection and extraction, graph connectivity, Quantum Cloud (Q-Cloud), Quantum Internet (Q-Int), Quantum Internet of Things (Q-IoT), image processing, healthcare applications, and more.

It is also possible to use other objective functions, such as entropy, encryption-decryption, optimization, feature selection, extraction, etc., to enhance the algorithm's thresholding performance. Furthermore, we can use trial and error to choose more parameter values for this experiment. Future studies should also focus on the adaptive and systematic adjustment of the D-QIGA settings.

## CRediT authorship contribution statement

**Akhilesh Kumar Singh:** Writing – review & editing, Writing – original draft, Visualization, Validation, Software, Resources, Methodology, Investigation, Formal analysis, Data curation, Conceptualization.

**Kirankumar R. Hiremath:** Review & editing, Validation, Supervision, Methodology, Investigation, Formal analysis, Data curation, Conceptualization.

**Conflict Of Interest**

Authors declare no conflict of interest

**Data Availability Statement**

Data will be made available on request.